\documentclass{article}

\usepackage{authblk}
\usepackage{adjustbox}

\usepackage{natbib}
\renewcommand{\bibname}{References}
\renewcommand{\bibsection}{\subsubsection*{\bibname}}

\usepackage{tikz}
\usetikzlibrary{positioning}
  \usetikzlibrary{tikzmark} 
  \usetikzlibrary{arrows}   
  \usetikzlibrary{calc}     
  \tikzstyle{every picture}+=[remember picture]
  \usetikzlibrary{shapes}
\usepackage{enumitem}

\usepackage[utf8]{inputenc} 
\usepackage[T1]{fontenc}    
\usepackage{csquotes}
\usepackage{hyperref}       
\usepackage{url}            
\usepackage{booktabs}       
\usepackage{amsfonts}       
\usepackage{nicefrac}       
\usepackage{microtype}      

\usepackage{amsmath}
\usepackage{amssymb}
\usepackage{tensor}
\usepackage{dsfont}

\usepackage{algorithm}
\usepackage{algpseudocode} 

\usepackage{appendix}
\usepackage{multibib}

\usepackage{graphicx}
\usepackage{subcaption}
\usepackage{wrapfig}

\usepackage{smartdiagram}
\hypersetup{
    colorlinks,
    linkcolor={red!50!black},
    citecolor={blue!50!black},
    urlcolor={blue!80!black}
}

\hypersetup{final}

\newtheorem{definition}{Definition}

\newcommand*{\matlab}{\textsc{Matlab}}
\newcommand*{\bvpfc}{\texttt{bvp5c}}

\newcommand{\bs}[1]{\boldsymbol{#1}} 
\renewcommand{\b}[1]{\mathbf{#1}} 
\newcommand{\Trans}{^{\intercal}} 
\newcommand{\vectorize}[1]{\text{vec}\!\left[ #1\right]} 
\newcommand{\parder}[2]{\frac{\partial#1}{\partial#2}} 

\def\M{\mathcal{M}} 

\newcommand*{\GP}{\mathcal{GP}}

\newcommand*{\N}{\mathcal{N}} 

\newcommand{\grad}[1]{\nabla_{#1}} 
\newcommand{\norm}[1]{\left\lVert#1\right\rVert} 
\newcommand*{\Id}{\mathbb{I}} 
\newcommand{\abs}[1]{\left\vert #1\right\vert} 

\newcommand*{\mesh}{\Delta}
\newcommand*{\bmesh}{\mathcal{B}}

\newcommand*{\data}{\mathcal{D}}
\newcommand*{\post}[1]{\underline{#1}}

\begin{document}

%

%


\title{Fast and Robust Shortest Paths on Manifolds Learned from Data}

%
%

\author[$\dagger$]{Georgios Arvanitidis}
\author[$\dagger$]{S{\o}ren Hauberg}
\author[$\ddagger$]{Philipp Hennig}
\author[$\star$]{Michael Schober}

\affil[$\dagger$]{Technical University of Denmark, Lyngby, Denmark}
\affil[$\ddagger$]{University of T\"ubingen,  T\"ubingen, Germany}
\affil[$\ddagger$]{Max Planck Institute for Intelligent Systems,  T\"ubingen, Germany}
\affil[$\star$]{Bosch Center for Artificial Intelligence, Renningen, Germany}


\maketitle

\begin{abstract}
We propose a fast, simple and robust algorithm for computing shortest paths and distances on Riemannian manifolds learned from data. This amounts to solving a system of ordinary differential equations (ODEs) subject to boundary conditions. Here standard solvers perform poorly because they require well-behaved Jacobians of the ODE, and usually, manifolds learned from data imply unstable and ill-conditioned Jacobians. Instead, we propose a fixed-point iteration scheme for solving the ODE that avoids Jacobians. This enhances the stability of the solver, while reduces the computational cost. In experiments involving both Riemannian metric learning and deep generative models we demonstrate significant improvements in speed and stability over both general-purpose state-of-the-art solvers as well as over specialized solvers.
\end{abstract}



\section{Introduction}

A long-standing goal in machine learning is to build models that are invariant
to irrelevant transformations of the data, as this can remove factors that
are otherwise arbitrarily determined. For instance, in nonlinear latent variable
models, the latent variables
are generally unidentifiable as the latent space is by design not invariant
to reparametrizations. Enforcing a Riemannian metric in the latent space that is invariant
to reparametrizations alleviate this identifiability issue, which significantly
boosts model performance and interpretability \citep{arvanitidis:iclr:2018, Tosi:UAI:2014}.
Irrelevant transformations of the data can alternatively be factored out by only modeling
local behavior of the data; geometrically this can be viewed as having a locally
adaptive inner product structure, which can be learned from data \citep{hauberg:nips:2012}.
In both examples, the data is studied under a Riemannian metric, so that can be seen as living on a Riemannian manifold.

\begin{figure}[t]
  \begin{center}
    \includegraphics[width=0.7\textwidth]{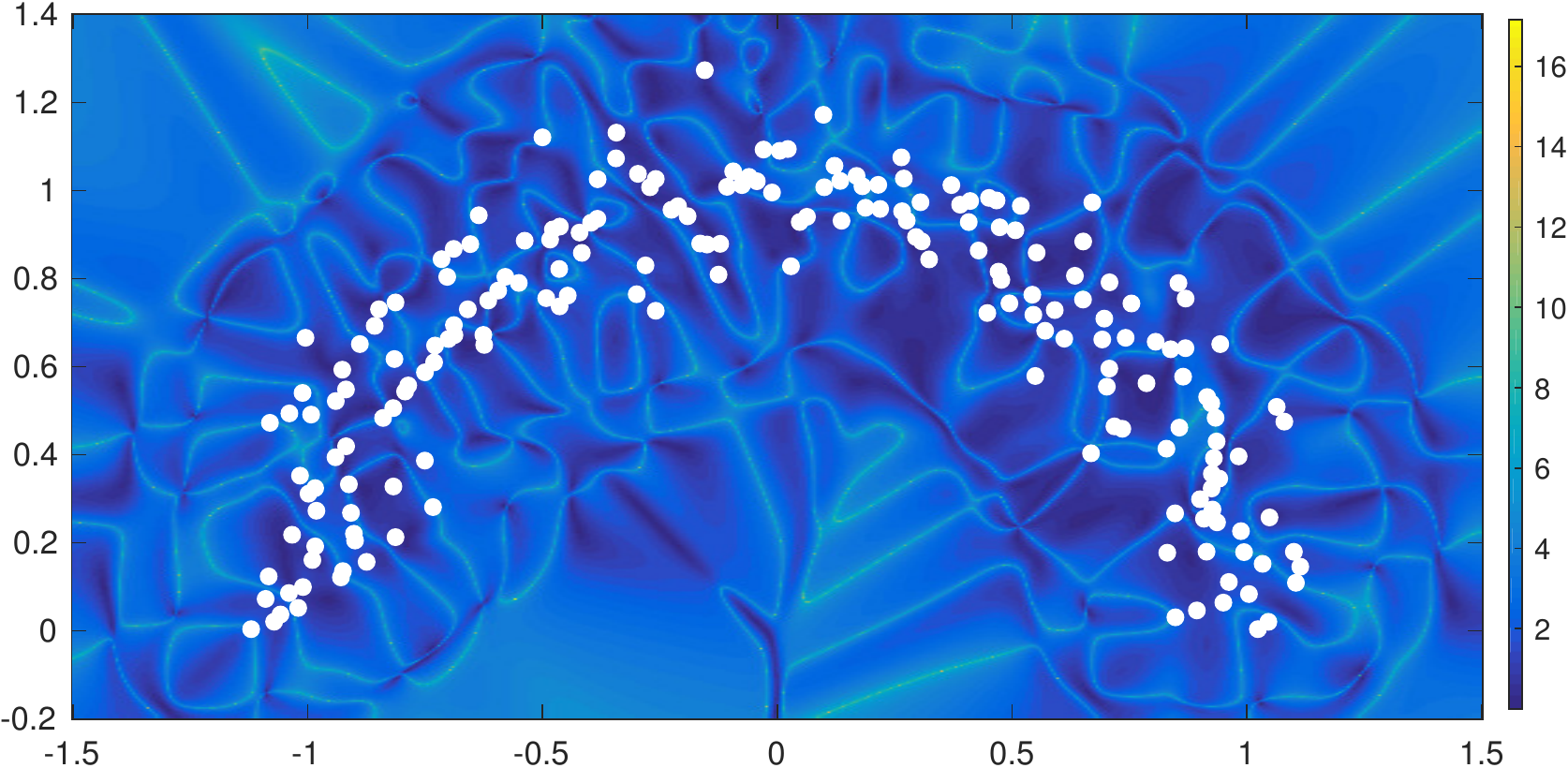}
  \end{center}
  \vspace{-10pt}
  \caption{A data manifold and the $\log$-condition number of the  Jacobian with fixed velocity (background).
  High condition numbers can cause failures to converge for off-the-shelf solvers.}
  \label{fig:jacobian_ill_conditioned}
\end{figure}

While this geometric view comes with strong mathematical support, it is
not commonly adopted. The primary concern is that the computational overhead
of Riemannian models often outweigh the induced benefits.
The main bottleneck herein are \emph{distance computations},
which are often at the core of machine learning algorithms.
In Euclidean space, distance evaluations require the computation of a norm.
In kernel methods, distances are evaluated via the kernel trick, but
many interesting geometries cannot be captured with
a positive definite kernel \citep{feragen2015geodesic}.
In both cases, distances can be computed with negligible effort.

The distance
between two points $\b{x}, \b{y}$ on a Riemannian manifold
is defined as the length of the \emph{shortest path}
between $\b{x}$ and $\b{y}$ known as the \emph{geodesic}.
Computing the geodesic requires the solution of a
\emph{boundary value problem (BVP)}.
These types of
\emph{ordinary differential equations (ODEs)}
 require specialized numerical methods for their solution,
as---unlike \emph{initial value problems (IVPs)}---they cannot be solved
via step-by-step algorithms like Runge--Kutta methods, and thus
their solution is computationally
\emph{taxing} \citep{ascher1994numerical}.

Moreover, Riemannian manifolds learned from data usually imply high curvature and an unstable Riemannian metric.
The reason is that the metric is estimated only from finite data,
so it changes irregularly fast.
As a consequence, the Jacobians of the ODE, which are required in many
off-the-shelf solvers, are often ill-conditioned which causes additional
problems and effort \citep[\textsection 8.1.2]{ascher1994numerical}.
The example in Fig.~\ref{fig:jacobian_ill_conditioned}
shows that the Jacobian associated with a data driven manifold is highly oscillatory. This implies that standard ODE solvers easily break down.
Thus, specialized BVP solvers for shortest path computations
are required to make distance evaluations on manifolds learned from data
as fast and robust as in other models.

In this paper, we propose a novel method for
computing shortest paths on Riemannian manifolds learned from data. Combining the theory of smoothing splines/Gaussian process
regression \citep{wahba1990spline,rasmussenwilliams} with
fixed-point iterations \citep{mann1953mean,johnson1972fixed},
we arrive at an algorithm which is \emph{simple}, \emph{fast} and
\emph{more reliable} on challenging geometries.
This is achieved by not utilizing the Jacobian
which is computationally costly and ill-behaved.

Our work is a significant improvement of an earlier algorithm
by \citet{HennigAISTATS2014}.
Their algorithm is an early proof of concept for
probabilistic numerics \citep{hennig15probabilistic}.
We show below that in the original form, it 
\emph{provably} does not converge to the true solution.
However, their algorithmic structure \emph{is capable} of
converging to the true solution.
We demonstrate \emph{how} their algorithm needs to be
adopted to improve \emph{solution quality} and
\emph{convergence speed}, so that the fixed-point
algorithm can efficiently compute the shortest path on Riemannian manifolds.


\section{A Brief Recap of Riemannian Geometry}

We start by defining \emph{Riemannian manifolds} \citep{docarmo:1992}. These are
well-studied metric spaces, where the inner product is only locally defined and changes
smoothly throughout space.

\begin{definition}
  A Riemannian manifold is a smooth manifold $\mathcal{M}$ where each tangent space
  $T_{\b{x}}\mathcal{M}$ is equipped with an inner product (Riemannian metric)
  $\langle \b{a}, \b{b} \rangle_\b{x} = \b{a}^{\Trans} \b{M}(\b{x}) \b{b}$
  that changes smoothly across the manifold.
\end{definition}

This inner product structure is sufficient for defining the length of a smooth curve
$\b{c}: [0, 1] \rightarrow \mathcal{M}$ in the usual way as
\begin{align}
  \mathrm{Length}(\b{c})
    &= \int_0^1 \sqrt{\dot{\b{c}}(t)^{\Trans} \b{M}(\b{c}(t)) \dot{\b{c}}(t) } \mathrm{d}t,
\end{align}
where $\dot{\b{c}} = \partial_t \b{c}$ denotes the curve \emph{velocity}.
The length of the shortest path connecting two points then constitutes the natural
distance measure on $\mathcal{M}$. The shortest curve is known as the geodesic, and can be found through
the Euler-Lagrange equations to satisfy a system of 2\textsuperscript{nd} order ODEs \citep{arvanitidis:iclr:2018},
\begin{align}
\label{eq:ode}
   \ddot{\b{c}}(t )
    = \frac{-{\b{M}({\b{c}(t)})}^{-1}}{2}\bigg[2 (\mathbb{I}_D \otimes  \dot{\b{c}}(t)^{\Trans})& \parder{\vectorize{\b{M}({\b{c}(t)})}}{\b{c}(t)}\dot{\b{c}}(t) \nonumber \\
     - \parder{\vectorize{\b{M}({\b{c}(t)})}}{\b{c}(t)}^{\Trans} &(\dot{\b{c}}(t) \otimes \dot{\b{c}}(t))\bigg],
\end{align}
where $\vectorize{\cdot}$ stacks the columns of a matrix into a vector and
$\otimes$ is the Kronecker product. 

Most numerical calculations on Riemannian manifolds are performed in local
tangent spaces as these are Euclidean. Key operations are therefore mappings
back and forth between the manifold and its tangent spaces. A point $\b{y} \in \mathcal{M}$
can be mapped to the tangent space at $\b{x}\in\mathcal{M}$ by computing the shortest connecting
curve and evaluating its velocity $\b{v}$ at $\b{x}$. This is a tangent vector at $\b{x}$
with the property that its length equals the length of the shortest path.
By the Picard-Lindel{\"o}f theorem \citep{picard1890}, this process can be reversed 
by solving Eq.~\ref{eq:ode} with initial conditions $\b{c}(0) = \b{x}$ and 
$\dot{\b{c}}(0) = \b{v}$. Mapping from the manifold to a tangent space, thus,
requires solving a boundary value problem, while the inverse mapping is an
initial value problem. Practically, the BVPs dominate the computational
budget of numerical calculations on manifolds.

\begin{figure}
  \begin{center}
    \includegraphics[width=0.5\textwidth]{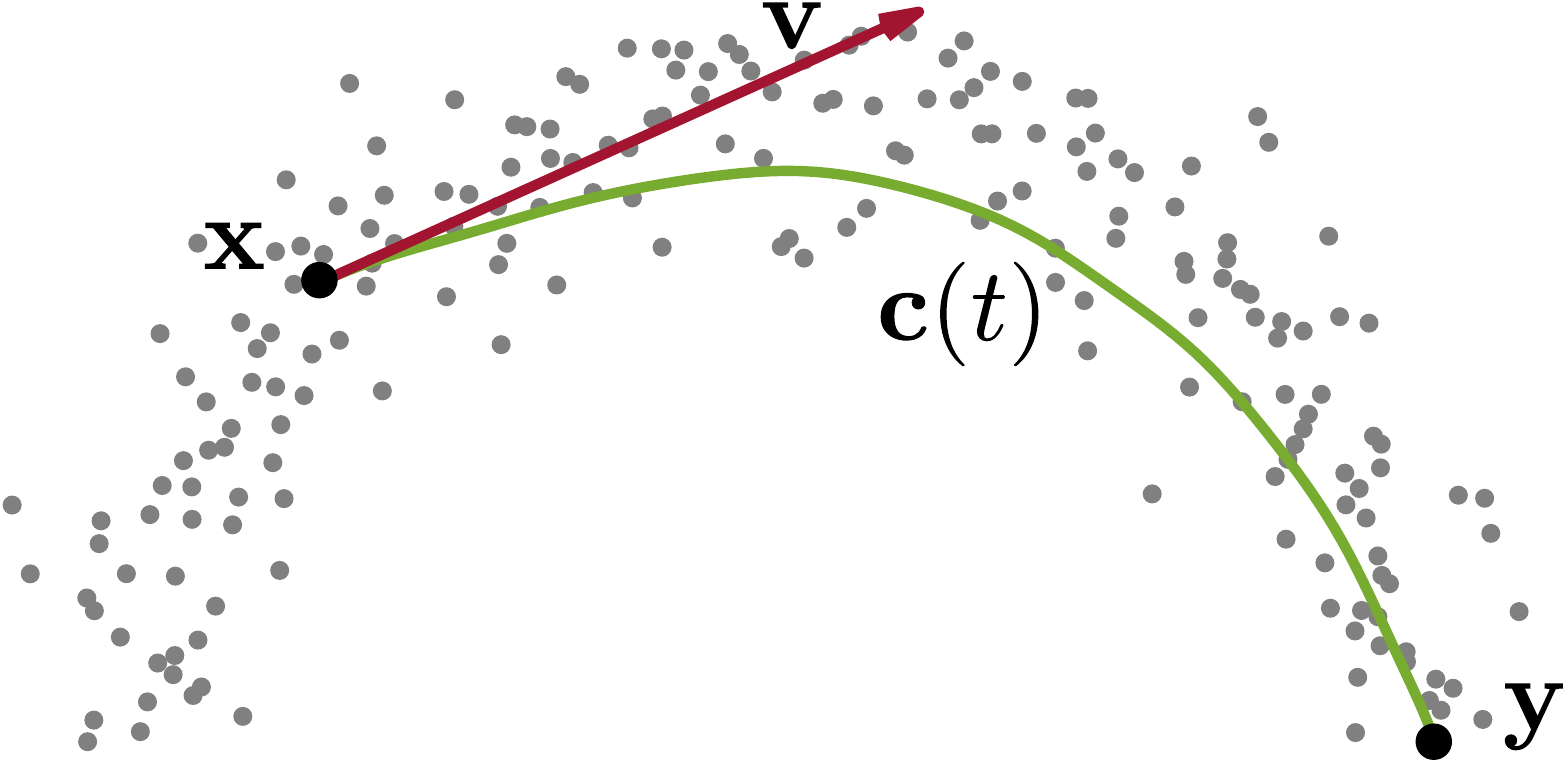}
  \end{center}
    \vspace{-10pt}
  \caption{A data manifold with a geodesic and its tangent vector.}
  \label{fig:teaser}
\end{figure}

In the context of this paper we will focus upon Riemannian manifolds which are \emph{learned} from the data, and capture its underlying geometric structure. Thus, we will consider for the smooth manifold the Euclidean space as $\mathcal{M} = \mathbb{R}^D$, and learn a Riemannian metric $\b{M}:\mathbb{R}^D \rightarrow \mathbb{R}^{D\times D}$. To be clear, this simply changes the way we measure distances, while respecting the structure of the data. An illustrative example can be seen in Fig.~\ref{fig:teaser}.
\newline


\section{A Fast Fixed-Point Method for Shortest Paths}

In order to apply Riemannian models to interesting data sets,
we require a fast and robust method to solve the boundary value problem 
\begin{equation}
\begin{aligned}
\ddot{\b{c}}(t) &= f(\b{c}(t),\dot{\b{c}}(t)),& \b{c}(0) = \b{x},&\;\b{c}(1) = \b{y}
\end{aligned}
\end{equation}
where $f(\b{c}(t),\dot{\b{c}}(t))$ is the right-hand side of Eq.~\eqref{eq:ode}
and $\b{x},\,\b{y} \in \mathcal{M}$.
Numerical BVP solvers typically replace the analytic
solution $\b{c}(t)$ by 
an approximant $\post{\b{c}}(t)$ which is required to
fulfill the ODE
$\post{\ddot{\b{c}}}(t_n) = f(\post{\b{c}}(t_n),\post{\dot{\b{c}}}(t_n))$
on a discrete \emph{mesh}
$\mesh = \{t_0 = 0, t_1, \dotsc, t_{N-1} = 1\} \subset [0, 1]$
of evaluation \emph{knots} $t_n$.
Together with the boundary conditions (BC)
$\post{\b{c}}(0) = \b{x},\;\post{\b{c}}(1) = \b{y}$,
this results in a $D(N+2)$-dimensional nonlinear
equation which can be solved for rich enough
$D(N+2)$-dimensional parametric models.
If the approximant is represented by the posterior
mean $\bs{\mu}(t)$ of a GP regressor
$\GP(\tilde{\b{c}}(t); \bs{\mu}(t), \b{k}(t,s))$, then
it can be proven that a solution with small
approximation error exists \citep[\textsection 11]{wendland2004scattered}
under suitable conditions on the kernel
\citep{micchelli2006universal,rasmussenwilliams}.
This means we have to find a (artificial) data
set $\data$ with $(N+2)$ $D$-dimensional data points
such that the posterior mean fulfills the BVP on
the discretization mesh $\mesh$ and the BC on
the curve \emph{boundary}
$\bmesh = \{0, 1\}$.
The data set $\data$ should be thought of as a
\emph{parametrization} for the solution curve.

One way to generate a solution is to define the
auxiliary function $F(\b{c}) = \ddot{\b{c}} - f(\b{c},\dot{\b{c}})$
and apply a variant of the Newton-Raphson method
\citep{deuflhard2011newton} to find a root of $F$.
For example, this type of algorithm is also at the core of
\matlab{}'s \bvpfc, a state-of-the-art BVP solver.
The convergence of these algorithms depends on
the Jacobians $\grad{\b{c}} f, \grad{\dot{\b{c}}} f$ of $f$
at the evaluation knots $t_n$.
In particular, Jacobians with big condition number
may cause Newton's method to fail to converge if
no precautions are taken \citep[\textsection 8.1.2]{ascher1994numerical}.
On manifolds learned from data, this is a common problem.
Furthermore, in practice these Jacobians are computed with
a computationally taxing finite difference scheme.
Thus, a method \emph{not} based on Newton's
method should be more suitable for the computation
of shortest paths.


\subsection{Method Description}
\label{sec:method_description}

As mentioned above, we model the approximate
solution $\post{\b{c}}(t)$ with the posterior mean
$\bs{\mu}(t)$ of a (multi-output) \emph{Gaussian process}
\begin{equation}
\GP(\tilde{\b{c}}(t);\,\bs{\mu}(t), \b{V} \otimes k(t,s))
\end{equation}
with (spatial) kernel $k$ and inter-dimensional covariance
matrix $\b{V} \in \mathbb{R}^{D\times D}$.
If the kernel $k$ is sufficiently partially differentiable,
this implies a covariance between derivatives of $\tilde{\b{c}}$
as well \citep[\textsection~4.1.1]{rasmussenwilliams}, in particular
\begin{equation}
\operatorname{cov}\left(\frac{\mathrm{d}^m}{\mathrm{d}t^m}\tilde{{c}}_i(t),
\frac{\mathrm{d}^n}{\mathrm{d}s^n}\tilde{{c}}_j(s)\right) = \b{V}_{ij} \frac{\partial^{m+n}}{\partial t^m \partial s^n} k(t,s)
\end{equation}
for the covariance between output dimensions $i$ and $j$, derivatives
$m$ and $n$ and spatial inputs $t$ and $s$.
This \emph{(prior) model} class is the same as in \citet{HennigAISTATS2014}.

The boundary equations fix two parameters $(0,\b{x}), (1, \b{y})$
of the parametrization. The remaining $N$ parameters
$(t_n, \ddot{\b{z}}_n)$ approximate the
\emph{accelerations} $\ddot{\b{c}}(t_n)$ of the true
solution $\b{c}(t_n)$ at knot $t_n$,
i.e., $\ddot{\b{z}}_n \approx \ddot{\b{c}}(t_n)$.
The $\ddot{\b{z}}_n$ are updated iteratively and
we denote values at the $i$-th iteration with the
superscript $(i)$, e.g., $\post{\b{c}}^{(i)}(t)$ for
the $i$-th approximation, $\ddot{\b{z}}_n^{(i)}$ for the
$i$-th value of the parameter $\ddot{\b{z}}_n$ and so forth.

At iteration $i$,
the approximation $\post{\b{c}}^{(i)}(t)$
is the predictive posterior $\GP$
\begin{equation}
  \begin{split}
  &P(\tilde{\b{c}}^{(i)}(t)) = \GP(\tilde{\b{c}}^{(i)}(t);\;\bs{\mu}^{(i)}(t), \b{k}^{(i)}(t,s))\\
  &\b{G} = \b{V} \otimes \bigg(\begin{bmatrix}k(\bmesh,\bmesh) & \frac{\partial^2}{\partial s^2}k(\bmesh, \mesh) \\ \frac{\partial^2}{\partial t^2}k(\mesh,\bmesh) &  \frac{\partial^4}{\partial t^2\partial s^2}k(\mesh,\mesh)\end{bmatrix} \\
    &\phantom{\b{G} = \b{V} \otimes \bigg(}
     + \operatorname{diag}(0,0,\b{\Sigma},\dotsc,\b{\Sigma})\bigg)\\
  &\bs{\omega}\Trans = \left(\b{V} \otimes
  \begin{bmatrix}k(t,\bmesh) & \frac{\partial^2}{\partial s^2}k(t,\mesh)\end{bmatrix}
  \right) \b{G}^{-1}\\
  &\bs{\mu}^{(i)}(t) = \b{m}(t) 
   + \bs{\omega}\Trans \text{vec}\left(
     \begin{bmatrix}\b{x} - \b{m}(0)\\ \b{y} - \b{m}(1)\\\ddot{\b{z}}^{(i)}_\mesh - \b{\ddot{m}}(\mesh)\end{bmatrix}\Trans \right)\\
  &\b{k}^{(i)}(t,s) = \b{V}\otimes k(t,s) - \bs{\omega}\Trans
     \left(\b{V} \otimes \begin{bmatrix}
       k(\bmesh, s)\\ \frac{\partial^2}{\partial t^2}(\mesh,s)
     \end{bmatrix} \right),
  \end{split}
  \label{eq:posterior}
  \raisetag{100pt}
\end{equation}
and similarly for the velocity $\post{\dot{\b{c}}}(t)$ by forming the $\b{G}$ and $\bs{\omega}$ accordingly.
In Eq.~\eqref{eq:posterior}, $\ddot{\b{z}}_\mesh \in \mathbb{R}^{N \times D}$ represents the accelerations, and $k(\bmesh,\mesh)\in\mathbb{R}^{2\times N}$ is the
matrix of kernel evaluations at boundary points and evaluation knots and 
similar for $k(\bmesh,\bmesh), k(\mesh,\mesh)$ and $k(\mesh,\bmesh)$.
Finally, $\bs{\Sigma} = \varepsilon \Id_D$ is the identity matrix times a small
regularization parameter $\varepsilon \approx 10^{-10}$, so ${\ddot{\bs{\mu}}}^{(i)}(t_n)\rightarrow \ddot{\b{z}}^{(i)}_n$.
The rationale for its inclusion will be postponed to the end of
Sect.~\ref{sec:comp-to-aistats} as it will become more apparent in contrast
to the model of \citet{HennigAISTATS2014}. Details for the  components $k(\cdot,\cdot),~\b{m},~\Delta$ in Appendix~\ref{sec:app:approximate_sp}. Now, we proceed with the description of the algorithm.

\definecolor{my_red}{RGB}{160,23,50}
\definecolor{my_green}{RGB}{120,171,58}
\definecolor{my_blue}{RGB}{16,116,186}

  \vspace{\baselineskip}
\begin{figure*}[!ht]
\begin{subfigure}[b]{0.32\textwidth}
	\begin{tikzpicture}
      \node(n1)  {\includegraphics[width=\textwidth]{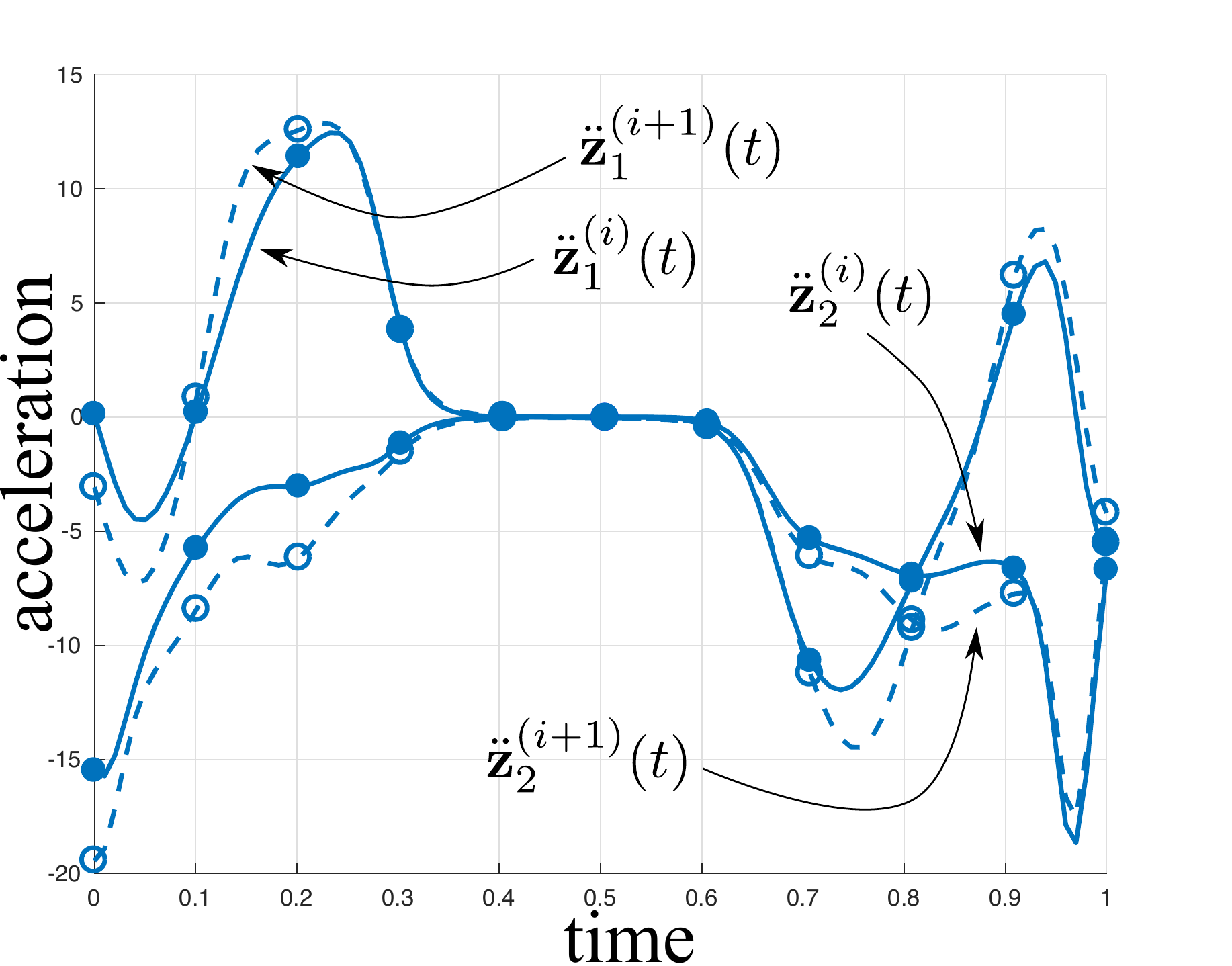}};
      \end{tikzpicture}
\end{subfigure}
 \begin{subfigure}[b]{0.32\textwidth}
 	\begin{tikzpicture}
       \node(n2) {\includegraphics[width=\textwidth]{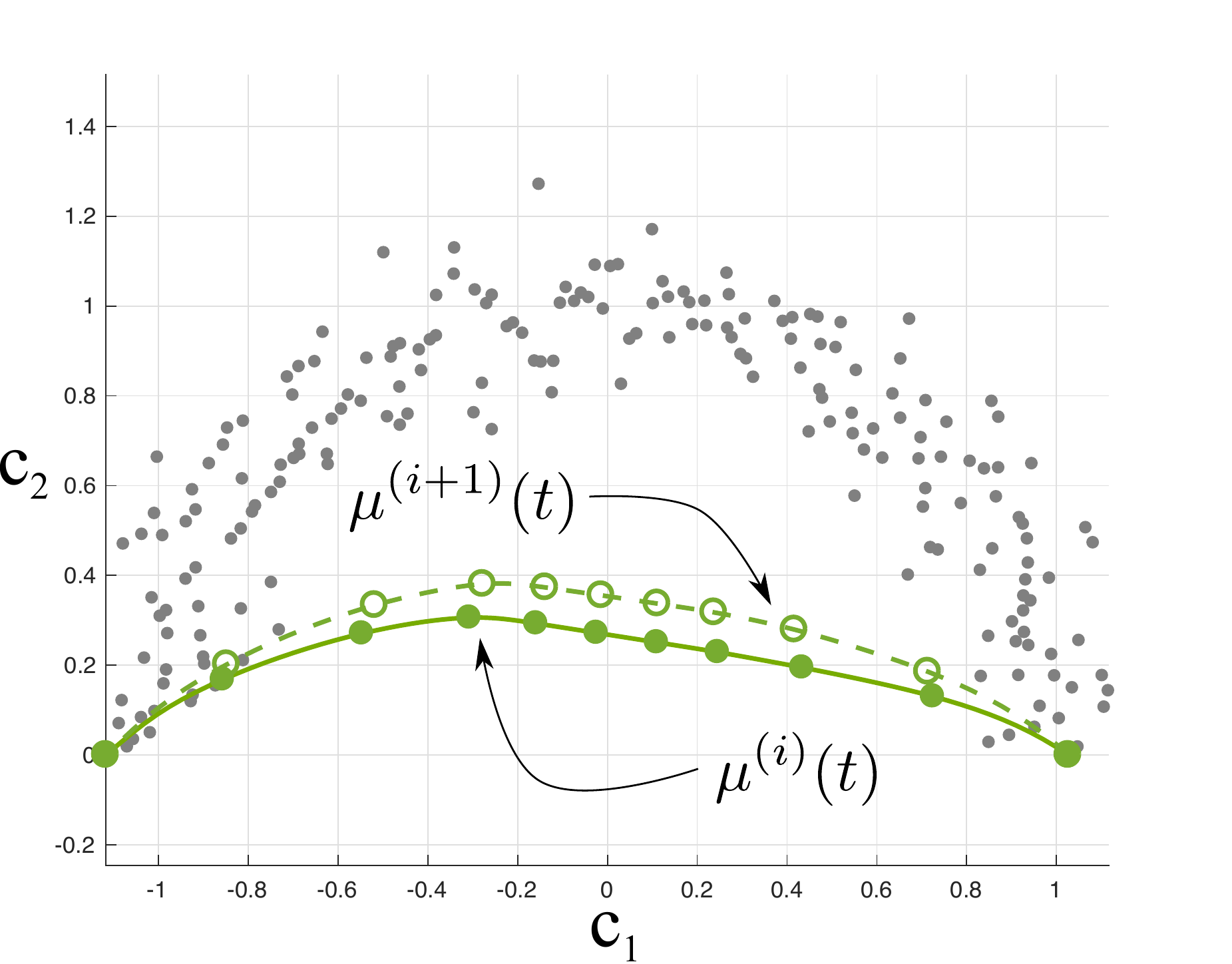}};
       \end{tikzpicture}
    \end{subfigure}
    \begin{subfigure}[b]{0.32\textwidth}
    \begin{tikzpicture}
        \node(n3){\includegraphics[width=\textwidth]{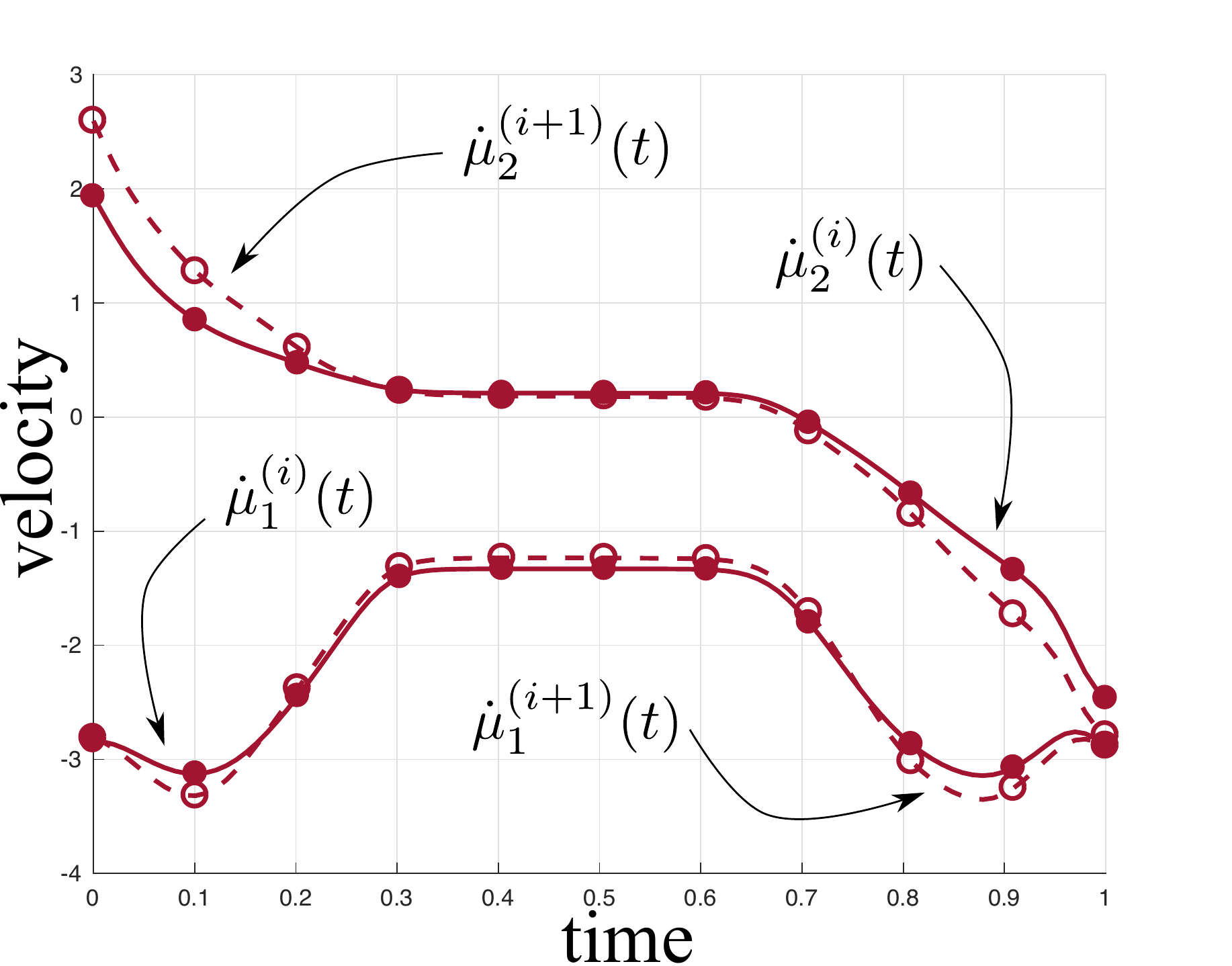}};
     \end{tikzpicture}
    \end{subfigure}
\begin{tikzpicture}[remember picture,overlay]
\draw [->,thick,black] ([xshift=0cm, yshift=1.5cm]n1.center) to[bend left=5] ([xshift=0cm, yshift=1.5cm]n2.center);
\draw [->,thick,black] ([xshift=0cm, yshift=1.5cm]n1.center) to[bend left=5] ([xshift=0cm, yshift=1.5cm]n3.center);
\node [fill=white,rounded corners=2pt] at ([xshift=+3cm, yshift=2.1cm]n1.center) {\textit{\textbf{Step 1}}};
\draw [->,thick,black] ([xshift=-0.5cm, yshift=-1.5cm]n2.center) to[bend left=5] ([xshift=1cm, yshift=-1.5cm]n1.center);
\draw [-,thick,black] ([xshift=-0.75cm, yshift=-1.5cm]n3.center) to[bend left=5] ([xshift=1cm, yshift=-1.5cm]n1.center);
\node [fill=white,rounded corners=2pt] at ([xshift=+3cm, yshift=-2.1cm]n1.center) {\textit{\textbf{Step 2}}};
\end{tikzpicture}
  \vspace{\baselineskip}
\caption[none]{\textbf{\emph{Step 1}}: With the current estimates of $\ddot{\b{z}}_\mesh^{(i)}$ (\begin{tikzpicture}
 {\draw [my_blue, line width=1.5] (0,0) -- (.4,0);}
 {\draw[my_blue, fill=my_blue] (0.2,0)  circle[radius=3pt];}
 \end{tikzpicture}) we generate using the $\GP$ model Eq. \ref{eq:posterior}, the current posterior curve $\bs{\mu}^{(i)}$ 
 (\begin{tikzpicture}
 {\draw [my_green, line width=1.5] (0,0) -- (.4,0);}
  {\draw[my_green, fill=my_green] (0.2,0)  circle[radius=3pt];}
 \end{tikzpicture}) 
 and velocities $\dot{\bs{\mu}}^{(i)}$ 
 (\begin{tikzpicture}
 {\draw [my_red, line width=1.5] (0,0) -- (.4,0);}
  {\draw[my_red, fill=my_red] (0.2,0)  circle[radius=3pt];}
 \end{tikzpicture}). \textbf{\emph{Step 2}}: Then, using the proposed fixed-point update scheme Eq. \ref{eq:fixed-point-update}, we get the updated parameters $\ddot{\b{z}}_\mesh^{(i+1)}$
 $(\begin{tikzpicture}
 {\draw [my_blue, densely dotted, line width=1.5] (0,0) -- (.4,0);}
{\draw[my_blue, line width=1] (0.2,0)  circle[radius=3pt];}
 \end{tikzpicture})$
. The algorithm iterates until $\norm{
\begin{tikzpicture}
{\draw [my_blue, densely dotted, line width=1.5] (0,0) -- (.4,0);}
 {\draw[my_blue, line width=1] (0.2,0)  circle[radius=3pt];}
\end{tikzpicture} 
-
f(
\begin{tikzpicture}
{\draw [my_green, densely dotted,  line width=1.5] (0,0) -- (.4,0);}
 {\draw[my_green, line width=1] (0.2,0)  circle[radius=3pt];}
\end{tikzpicture}
,
\begin{tikzpicture}
{\draw [my_red, densely dotted, line width=1.5] (0,0) -- (.4,0);}
 {\draw[my_red, line width=1] (0.2,0)  circle[radius=3pt];}
\end{tikzpicture} 
)
 }$ small enough.}
\label{fig:algorithm}
\end{figure*}

Just like a root of the function
$F(\b{c}) = \ddot{\b{c}} - f(\b{c},\dot{\b{c}})$
solves the ODE, so does a fixed point of the mapping
$\ddot{\bs{\mu}}^{(i+1)}(t) = f(\bs{\mu}^{(i)}(t),\dot{\bs{\mu}}^{(i)}(t))$.
In particular, we can evaluate this mapping on the
discretization mesh $\mesh$ to map $\ddot{\b{z}}_\mesh^{(i)}$
to $\ddot{\b{z}}_\mesh^{(i+1)}$. 
The big advantage of this combination of parametrization and update scheme
is the simplicity of obtaining closed-form
iteration updates
\citep[\textsection~9.4]{rasmussenwilliams}.
The vector field $f$ is evaluated
using the current iteration
$(\bs{\mu}^{(i)},\dot{\bs{\mu}}^{(i)})$ to yield
$\ddot{\b{z}}_n^{(i+1)} = f(\bs{\mu}^{(i)}(t_n),\dot{\bs{\mu}}^{(i)}(t_n))$, and $\ddot{\b{z}}_n^{(i+1)} \approx \ddot{\bs{\mu}}^{(i+1)}(t_n)$ because $\varepsilon \rightarrow 0$.
Forming $\bs{\mu}^{(i+1)}, \dot{\bs{\mu}}^{(i+1)}$ from $\ddot{\b{z}}_n^{(i+1)}$
only requires two matrix-vector products (see Eq.~\eqref{eq:posterior}).
The process is depicted in Fig.~\ref{fig:algorithm}.

\begin{algorithm}[H]
  \caption{The proposed fixed-point method.}
  \label{alg:bvp-solver}
\begin{algorithmic}[1]
\Require {BVP $f(\b{c}(t),\dot{\b{c}}(t))$, hyper-parameters $\mesh, \varepsilon$, tolerance $\tau$}
\State \emph{\# Compute $\bs{\mu}^{(i)}(t), \dot{\bs{\mu}}^{(i)}(t)$ using Eq. \ref{eq:posterior} and $\ddot{\b{z}}^{(i)}_\mesh$.}
\State Define: $e^{(i)}_n \triangleq \norm{\ddot{\b{z}}^{(i)}_n - f(\bs{\mu}^{(i)}(t_n),            \dot{\bs{\mu}}^{(i)}(t_n))}^2$
\State $\ddot{\b{z}}_n^{(0)} \gets \b{0},~ n=0,\dots,N-1$
\State $i \gets 0$ 
\While{$\exists n:
  e^{(i)}_n
        > \tau$}
\State $\ddot{\b{z}}_n^* \gets f(\bs{\mu}^{(i)}(t_n), \dot{\bs{\mu}}^{(i)}(t_n)),~n=0,\dots,N-1$
\For{$j=0,\dotsc,3$}\label{line:backtrack-start}
\State $\alpha_j = 3^{-j}$
\State $\ddot{\b{z}}_{\mesh}^{(*,j)} \gets \alpha_j \ddot{\b{z}}_\mesh^* + (1 - \alpha_j) \ddot{\b{z}}_\mesh^{(i)}$
\If{$\sum_n e^{(*,j)}_n \leq \sum_n e^{(i)}_n$ }
\State \texttt{break}
\EndIf
\EndFor\label{line:backtrack-end}
\State $\ddot{\b{z}}_\mesh^{(i+1)} \gets \ddot{\b{z}}_{\mesh}^{(*,j)}$
\State $i \gets i+1$
\EndWhile
\State
\Return $\GP(\tilde{\b{c}}(t);\;\bs{\mu}^{(i)},\b{k}^{(i)})$
\end{algorithmic}
\end{algorithm}



Variants of this scheme have been repeatedly applied
for the creation of probabilistic
differential equation solvers
\citep{HennigAISTATS2014,chkrebtii2016bayesian,schober2014nips,cockayne2016probabilistic,KerstingHennigUAI2016,teymur2016probabilistic,schober2017probabilistic,kersting2018arXiv}.
Of these papers, only \citet{HennigAISTATS2014} points out that this can be
updated multiple times, but even there the connection between a fixed point
of the mapping and an approximate solution is not stated.
Interpreting the iteration as a fixed point search is the
key insight of this paper.


We suggest to apply a \emph{Mann iteration}
\citep{mann1953mean,johnson1972fixed} process
for the solution of \eqref{eq:ode} given by
\begin{equation}
\begin{aligned}
\label{eq:fixed-point-update}
\ddot{\b{z}}_n^* &= f(\bs{\mu}^{(i)}(t_n),\dot{\bs{\mu}}^{(i)}(t_n))\\
\ddot{\b{z}}_n^{(i+1)} &= \alpha_i \ddot{\b{z}}_n^* + (1 - \alpha_i) \ddot{\b{z}}_n^{(i)}
\end{aligned}
\end{equation}
with \enquote{step sizes} $\alpha_i \in [0, 1]$.
The results of \citet{mann1953mean,johnson1972fixed}
only apply if $\alpha_i = (i+1)^{-1}$, however we found
a backtracking scheme to be effective in practice.


Algorithm \ref{alg:bvp-solver} presents our method
in pseudo-code where $\ddot{\b{z}}_{\mesh}^{(*,j)}$ denotes the tentative parametrization. Note how the backtracking line search for $\alpha_i$
(Lines \ref{line:backtrack-start}-\ref{line:backtrack-end})
requires half of the description.

Our method is similar to a recently proposed
method by \citet{BELLO2017} that is based
on the \emph{variational iterative method}\footnote{\emph{calculus of variations} not \emph{variational inference}.} by \citet{HE2000115}.
\citet{BELLO2017} proposed to use this
scheme \emph{symbolically} requiring a computer-algebra system
for its execution, which makes it inapplicable to practical tasks.
More details to these related works can be found in
\citet{JAFARI20141,KHURI201428}.

\subsection{Comparison with \citet{HennigAISTATS2014}}
\label{sec:comp-to-aistats}

The proposed method is inspired by the previous work
of \citet{HennigAISTATS2014} and we make a direct comparison here.

The algorithm of \citet{HennigAISTATS2014} is a proof-of-concept
probabilistic numerical method \citep{hennig15probabilistic}
for solving boundary value problems.
It is structurally similar to other early probabilistic IVP
solvers of \citet{chkrebtii2016bayesian}
and \citet{skilling1991bayesian}.
Since the publication of these early works,
the field  has matured significantly,
providing algorithms with novel functionality
\citep{hauberg2015random,nips2015_5753,oates2017bayesian,briol2018bqmultiple}
and rigorous analysis
\citep{briol2015frank,chkrebtii2016bayesian,schober2017probabilistic,kersting2018arXiv}.

Their main idea is to treat the vector-field
evaluations $\ddot{\b{z}}_n$
as \emph{noisy observations} of the true, but unknown,
second derivative $\ddot{\b{c}}(t_n)$.
For a concrete suggestion, they propose a Gaussian
likelihood
$P(\ddot{\b{z}}_n) = 
\N(\ddot{\b{z}}_n; \ddot{\b{c}}(t_n), \b{\Lambda}_n)$.
Together with a GP prior on $\b{c}(t_n)$,
they arrive at an inference algorithm.
Heuristically, they propose to add mesh observations
sequentially, refine them iteratively for a fixed number
of steps, and they repeat the overall process until they
find a set of hyper-parameters of the GP which maximizes the
data likelihood of the final approximation.

However, the algorithm of \citet{HennigAISTATS2014}
cannot converge to the true solution in general.
A convergent method is required to satisfy
$f(\bs{\mu}^{(i)}(t_n),\dot{\bs{\mu}}^{(i)}(t_n)) \to \ddot{\bs{\mu}}^{(i)}(t_n)$
as $i \to \infty$.
However, as $\b{\Sigma}_n \neq \b{0}$ in
\citet{HennigAISTATS2014}, $\ddot{\bs{\mu}}^{(i)}(t_n)$ is
\emph{not} an interpolant of $\ddot{\b{z}}_n^{(i)}$
(\citeauthor{kimeldorf1970correspondence}, \citeyear{kimeldorf1970correspondence}, Thm.~3.2;
\citeauthor{kanagawa2018gps}, \citeyear{kanagawa2018gps}, Prop.~3.6)
implying that the \emph{true accelerations}
$\ddot{\b{c}}(t_n)$ cannot be a parameterization
in the model of \citet{HennigAISTATS2014}
contradicting the fixed point requirement.

The same criticism could be applied to our model,
as we propose $\b{\Sigma}_n = \b{\Sigma} = \varepsilon \Id_D \neq \b{0}$.
We have experimented with annealing schemes for
this hyper-parameter $\b{\Sigma}^{(i)} = i^{-1} \b{\Sigma}$,
but the benefit of $\varepsilon > 0$ for the stability of the
Gram matrix $\b{G}$ is bigger than induced numerical inprecision,
in particular when the tolerance $\tau$ is considerably larger
than $\varepsilon$.

Although the algorithm of \citet{HennigAISTATS2014} cannot converge,
three insights and resulting modifications lead to our proposed method:

\begin{enumerate}
  \item \citet{HennigAISTATS2014} did not propose a principled scheme to determine the number of refinements $S$, but treated
  it as a hyper-parameter that must be provided by the user.
  However, it can be easily checked whether the posterior
  mean $\bs{\mu}^{(i)}(t)$ fulfills the differential equation
  at any point $t$. This not only removes a hyper-parameter, but
  can also gives more confidence in the returned solution.
  In principle, the error
  $e_n^{(i)} = \norm{\ddot{\bs{\mu}}^{(i)}(t_n) 
  - f(\bs{\mu}^{(i)}(t_n),\dot{\bs{\mu}}^{(i)}(t_n))}$
  could even be used to construct adaptive meshes $\mesh^{(i)}$
  \citep{mazzia2004hybrid}.
  
  \item Using an universal kernel \citep{micchelli2006universal}
  and a fine enough mesh $\mesh^*$, it is known that any curve
  can be fitted.
  While sub-optimal kernel parameters $\theta$ might require
  an exponential bigger mesh
  \citep[Thm.~10]{vaart2011information}, the property
  of universality holds \emph{regardless} of the
  hyper-parameter $\theta$ used to find the approximation.
  As a consequence, tuning hyper-parameters is \emph{purely optional} and certainly \emph{does not require restarts}.
  This also improves runtime significantly as the Gram matrix
  needs only be to inverted \emph{once}.
  In practice, we have observed negligible solution improvements
  after type-II maximum likelihood optimization after the
  end of the algorithm.

  \item Currently, there is no analysis when or if
  coordinate-wise updates offer improvements over
  simultaneous updates for all parameters $\ddot{\b{z}}^{(i)}_\mesh$.
  There is, however, a strong argument \emph{for} updating
  simultaneously: runtime. All predictive posterior parameters
  can be pre-computed and kept fixed throughout the runtime
  of the algorithm, if the mesh is not adapted throughout.
  In particular, the regression weights $\bs{\omega}\Trans$
  can be kept fixed, so each update requires
  only two vector-vector products.
\end{enumerate}

Finally, while our derivation does not make use of its
probabilistic interpretation, further steps in this
direction could potentially unlock novel functionality
which has repeatedly been the case with other
probabilistic numerical methods
\citep{briol2018bqmultiple,oates2017bayesian,hauberg2015random}.


\section{Experiments}

\begin{figure*}[ht]
    \centering
    \begin{subfigure}[b]{0.31\textwidth}
        \includegraphics[width=\textwidth]{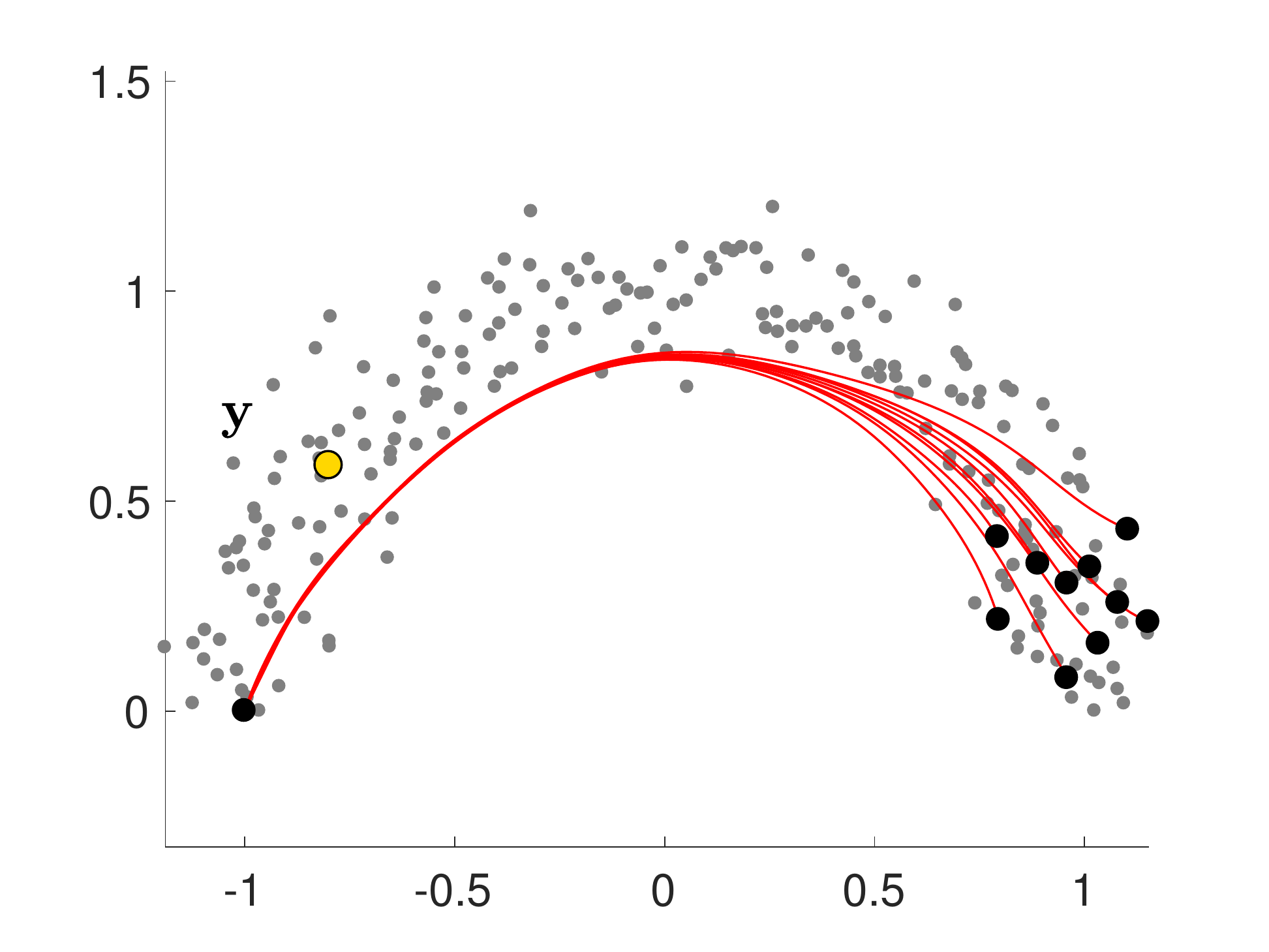}
        \caption{Data}
        \label{fig:ex1_fig1}
    \end{subfigure}
    ~ 
    \begin{subfigure}[b]{0.31\textwidth}
        \includegraphics[width=\textwidth]{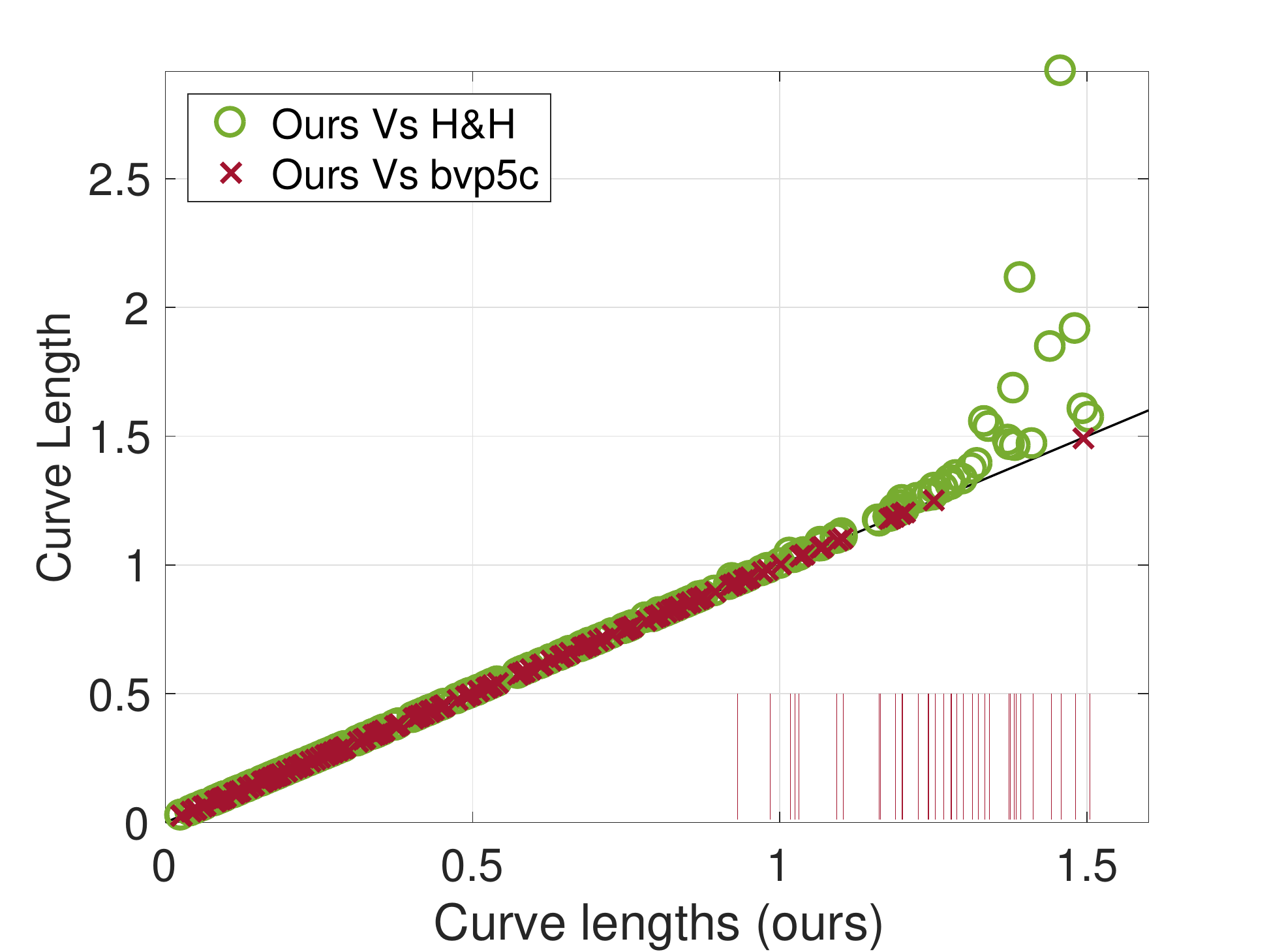}
        \caption{Curve Lengths}
        \label{fig:ex1_fig2}
    \end{subfigure}
    ~ 
    \begin{subfigure}[b]{0.31\textwidth}
        \includegraphics[width=\textwidth]{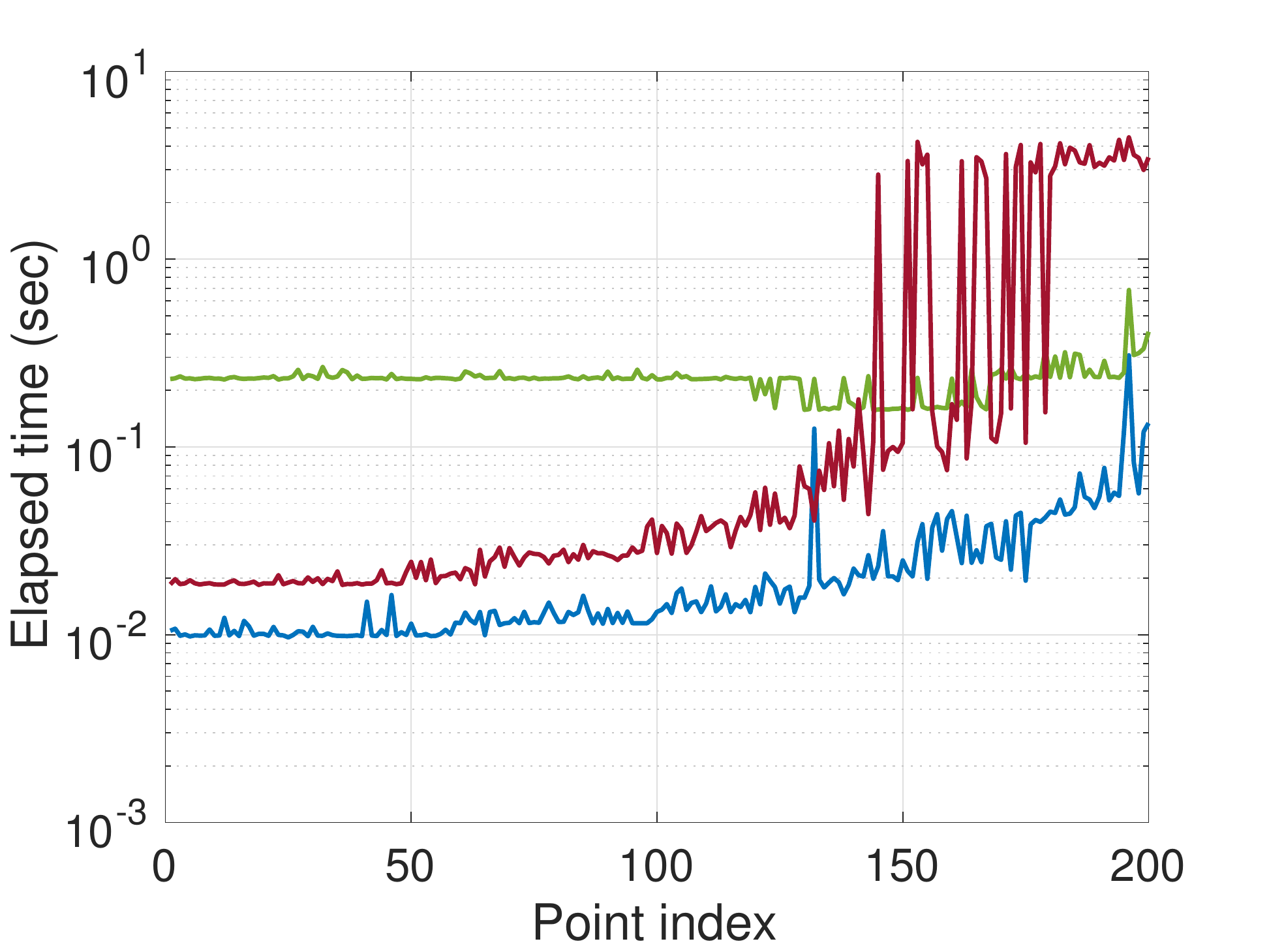}
        \caption{Runtimes }
        \label{fig:ex1_fig3}
    \end{subfigure}
    \caption{\textit{Left}: Generated data on a semi-circle, together with some challenging geodesics computed by our method, and a point $\mathbf{y}$. \textit{Middle}: The curve lengths of the geodesics between the given data and the point $\mathbf{y}$, on the horizontal axis we have the point index, and the vertical lines represent the failures of the \bvpfc{} to converge. \textit{Right}: The runtime for the corresponding geodesic problems.}\label{fig:ex1}
\end{figure*}

In this section we demonstrate the advantages of our method compared to \matlab{}'s \bvpfc{}, and the algorithm in \citet{HennigAISTATS2014} denoted in the experiments as H{\&}H. Since all the methods depend on a set of parameters, we will come up with a \emph{default} setting. 
For the proposed method, we will use for the $\mesh$ a uniform grid of $N=10$ points including the boundaries. The corresponding noise term of the points $\ddot{\b{z}}$ will be kept fixed to $\b{\Sigma} = 10^{-7}\mathbb{I}_D$. For the GP we will use the Squared Exponential kernel $k(t,t') = \exp(-(2\lambda^2)^{-1}\abs{t-t'}^2)$. We fix the amplitude $\b{V}$ in a Bayesian fashion as \cite{HennigAISTATS2014}, and the length-scale $\lambda^2 \approx 2^{-1} \abs{t_{n+1} - t_n}$ which provides enough degrees-of-freedom while covering the entire interval at the same time.
The prior mean is set to the straight line $\b{m}(t) = \b{c}(0) + t\cdot(\b{c}(1) - \b{c}(0))$, and the derivative accordingly.
For the method of \citet{HennigAISTATS2014}, we use the same
parameters as for our method.
We set the bounds on the Jacobian to $U = \dot{U} = 10$ and
run the method with one refinement iteration.
We set the maximum mesh size for the \bvpfc{} to $1000$, and use for the starting mesh uniformly $10$ points on the straight line connecting the boundary points. For all the methods we consider the resulting curve as correct if $\norm{\ddot{\b{c}}(t_n) - f(\b{c}(t_n), \dot{\b{c}}(t_n))}_2^2 \leq 0.1, ~\forall n$.

\subsection{Experiments with a Non-parametric Riemannian Metric}
\label{sec:non_parametric_experiments}

We first consider the case of Riemannian metric learning as proposed by \cite{arvanitidis:nips:2016}. This can be seen as a way to capture the local density of the data, and thus to uncover the underlying geometric structure. The metric at a given point is computed in three steps: 1) use a kernel to assign weight to given data, 2) compute the local diagonal covariance matrix, and 3) use its inverse as the metric tensor. More formally, the metric is
\begin{align}
\b{M}_{dd}(\b{x}) &= \left(\sum_{n=1}^N w_n(\b{x})(x_{nd} - x_d)^2 + \rho \right)^{-1}, \\ &\text{where} \quad w_n(\b{x}) = \exp\left( -\frac{\norm{\b{x}_n - \b{x}}^2_2}{2\sigma_{\mathcal{M}}^2}\right). \nonumber
\end{align}
The parameter $\sigma_{\mathcal{M}} \in \mathbb{R}$ controls the curvature of the manifold, since it regulates the rate of the metric change i.e. when the $\sigma_{\mathcal{M}}$ is small then the metric changes fast so the curvature increases. The parameter $\rho\in \mathbb{R}_{>0}$ is fixed such that to prevent zero diagonal elements.

We generated 200 data points along the upper semi-circle and added Gaussian noise $\mathcal{N}(0,0.01)$ as it is shown in Fig. \ref{fig:ex1_fig1}. The $\rho=0.01$ is kept fixed in all the experiments. We set the parameter $\sigma_{\mathcal{M}}=0.15$, and after fixing the point $\b{c}(0)=\b{y}$ we compute the geodesics to the given data. From the results (Fig. \ref{fig:ex1_fig2}) we see that all three methods perform well when the distance from the starting point is small. However, when distances increase, we see that only our method manages to find the correct curve, while \bvpfc{} is not able to solve the problem, and H{\&}H finds too long curves. Also, we see that the runtime of our method (Fig. \ref{fig:ex1_fig3}) is increased only slightly for the difficult problems, while \bvpfc{} is always slower and especially for the difficult problems increases the mesh size to the maximum and fails to converge. The performance of the H{\&}H remains almost constant, since it is essentially based on the converge of the model's parameters and not the difficulty of the problem.

\begin{figure}[ht]
  \begin{center}
    \includegraphics[width=0.75\textwidth]{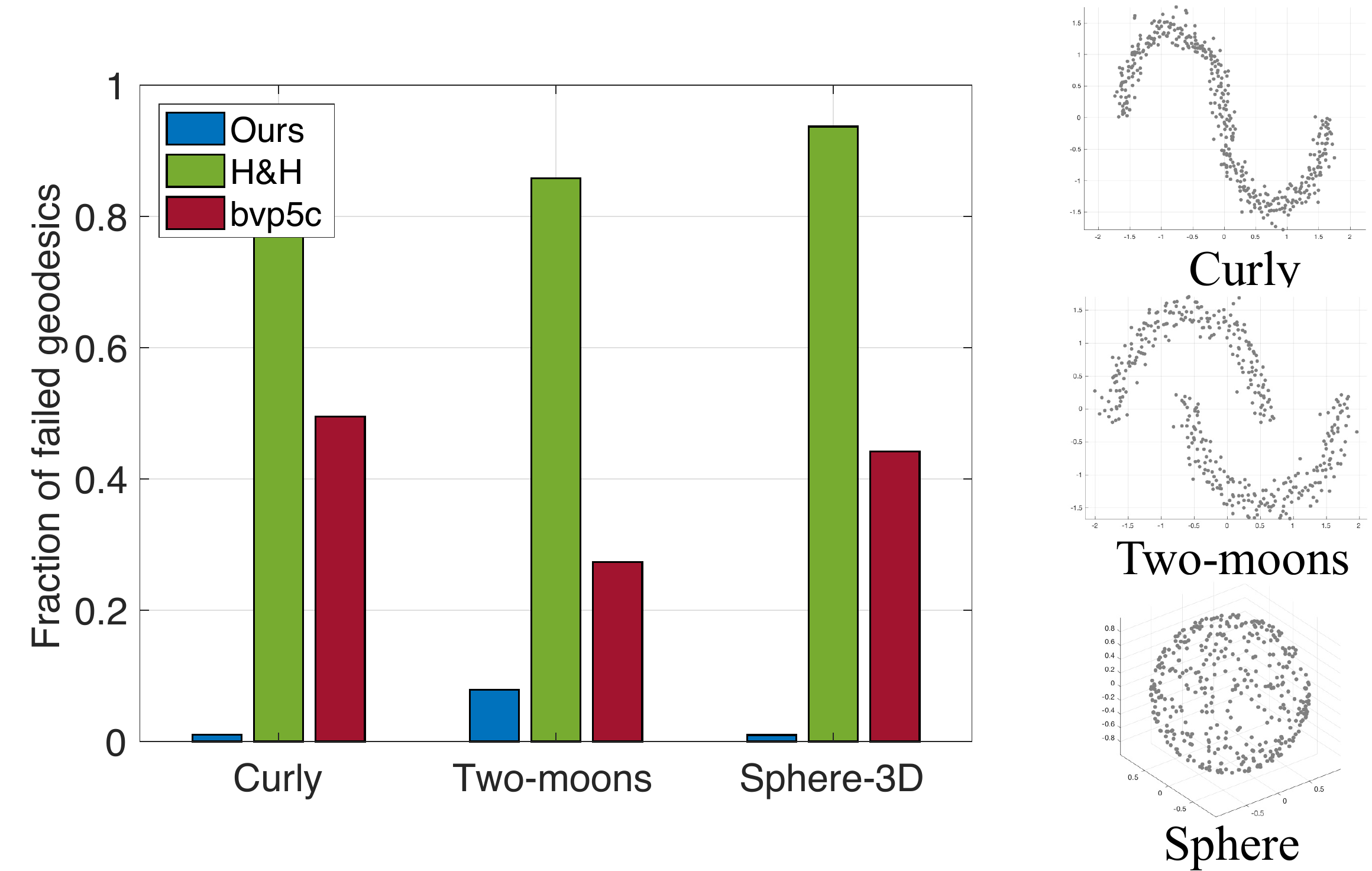}
  \end{center}
  \vspace{-10pt}
  \caption{Failed geodesics.}
  \label{fig:ex2}
\end{figure}
Next, we generated three challenging datasets consisting of 400 points each. For the first one we generate a circle and we flip the lower half along the $y$ axis, we refer to it as Curly in the results. For the second we move the lower half of the circle such that to get the two moons. The third one is a 2-dimensional sphere in $\mathbb{R}^3$. Finally, we add Gaussian noise $\mathcal{N}(0,0.01)$, and we standardize to zero mean and unit variance each dimension. We keep the same parameters for the metric. Note that the resulting manifold implies high curvature, so we increased the flexibility of the methods. For the proposed model and H{\&}H we used a grid of 50 points, and the maximum mesh size of \bvpfc{} was set to 5000. We pick randomly 40 points for each dataset, and compute the pairwise distances. In Fig. \ref{fig:ex2} we see that our method manages to solve almost all of the shortest path problems, while \bvpfc{} fails in almost half of them. H{\&}H is expected to fail in many cases, since it is not designed to converge to the correct solution.


\begin{figure}[ht]
\centering
    \includegraphics[width=0.5\textwidth]{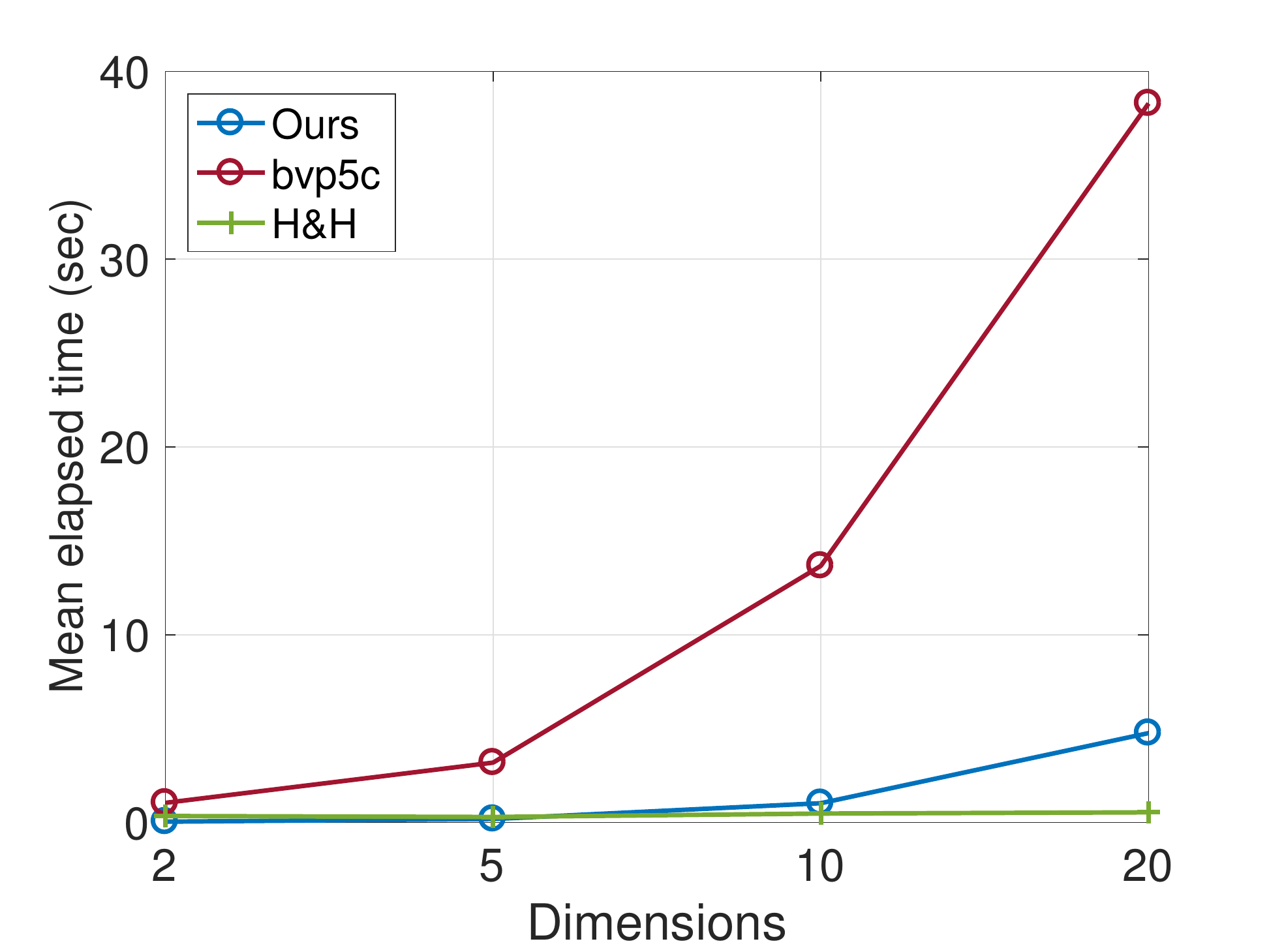}
    
     \begin{tabular}{c | c c c c }
		D & 2 & 5 & 10  & 20\\\hline
		Ours & 0\% & 0\% & 0\% &  9\%\\
		\bvpfc{} & 9\% & 24\% & 42\% & 50\%\\
		H{\&}H & 69\% & 95\% & 100\% & 100\%
	\end{tabular}
  \vspace{-5pt}
  \caption{Scalability in higher dimensions and failures.}
  \label{fig:ex3}
\end{figure}
Furthermore, we tested the scalability of the methods with respect to dimensionality. We generate 1000 points on a semi-circle in 2 dimensions, and standardized to zero mean and unit variance. We fix a point $\b{y}$ and a subset $\mathcal{S}$ of 100 points. Then, for every $D = [2, 5, 10, 20]$ we construct an orthogonal basis to project the data in $\mathbb{R}^D$, where we standardize the data again and add Gaussian noise $\mathcal{N}(0,0.01)$. Then, for each dimension $D$ we compute the geodesics between the $\b{y}$ and the subset $\mathcal{S}$ of the points. Keep in mind that the parameter $\sigma_{\mathcal{M}}=0.25$ is kept fixed, so as the dimensions increase the sparsity of the data increase, and so does the curvature of the manifold. In Fig.~\ref{fig:ex3} we show the average runtime for every dimensionality and the percentage of failures for each method. Our method remains fast in higher dimensions, even if the curvature of the manifold is increased. On the other hand, we observe that \bvpfc{} fails more often, which causes the overhead in the runtime. H{\&}H remains fast, but, the resulting curves cannot be trusted as the criterion of correct solution is almost never met.

\begin{figure}[ht]
  \begin{center}
    \includegraphics[width=0.5\textwidth]{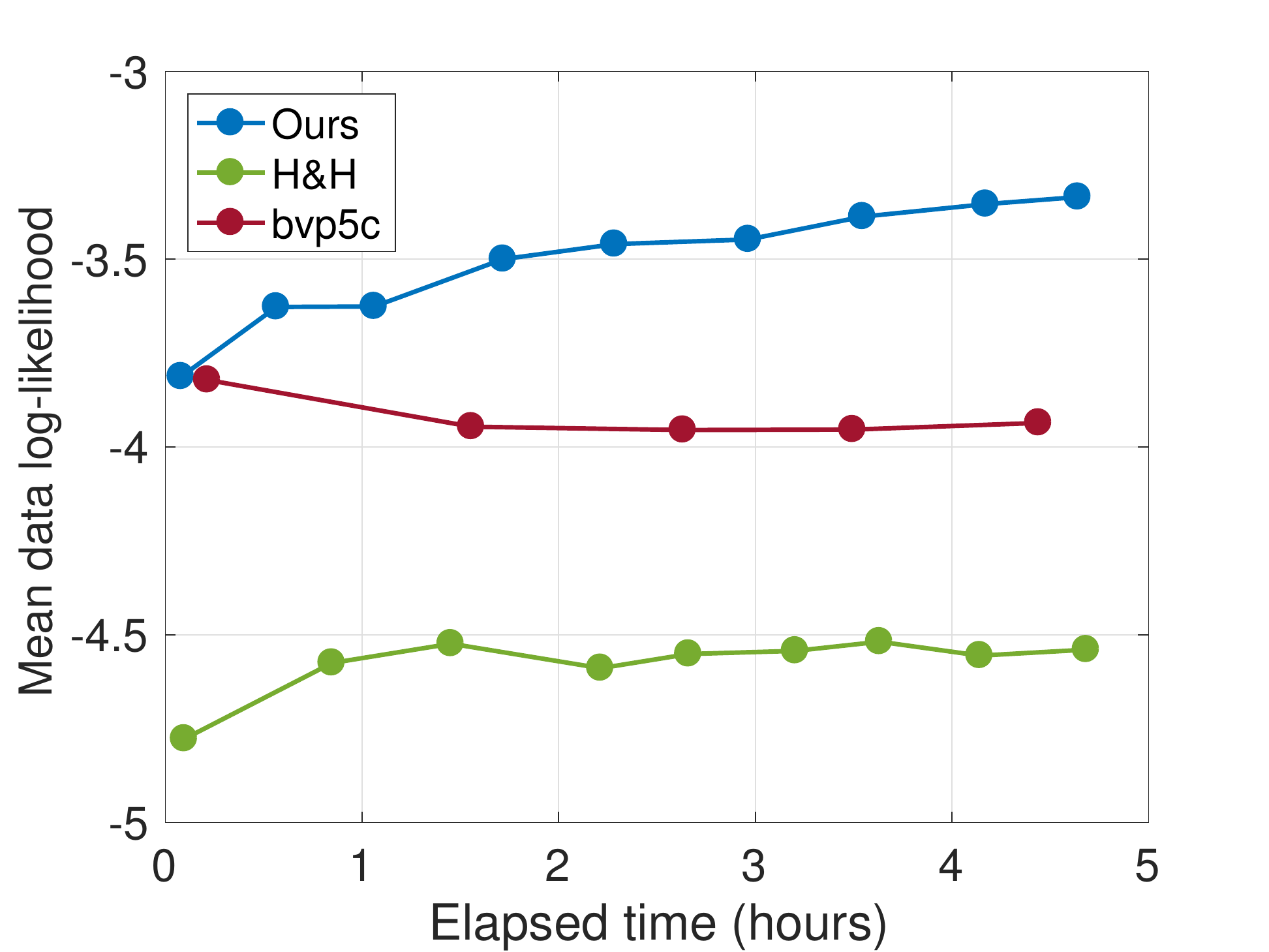}
  \end{center}
  \vspace{-10pt}
  \caption{LAND experiment.}
  \label{fig:ex8}
\end{figure}
Also, we fitted a mixture of LANDs \citep{arvanitidis:nips:2016} using the three models, on the two moons dataset generated by flipping and translating the data in Fig.~\ref{fig:ex1_fig1}. Note that we fix $\sigma_{\mathcal{M}}=0.1$ since we want our metric to capture precisely the underlying structure of the data, which implies that the curvature is increased.  From the results in Fig.~\ref{fig:ex8} we see that the proposed solver faster achieves higher log-likelihood. Additionally, in the same time interval, it manages to run more iterations (dots in the figure).

\subsection{Experiments with a Riemannian Metric of Deep Generative Models}

\begin{figure*}[ht]
    \centering
    \begin{subfigure}[b]{0.31\textwidth}
        \includegraphics[width=\textwidth]{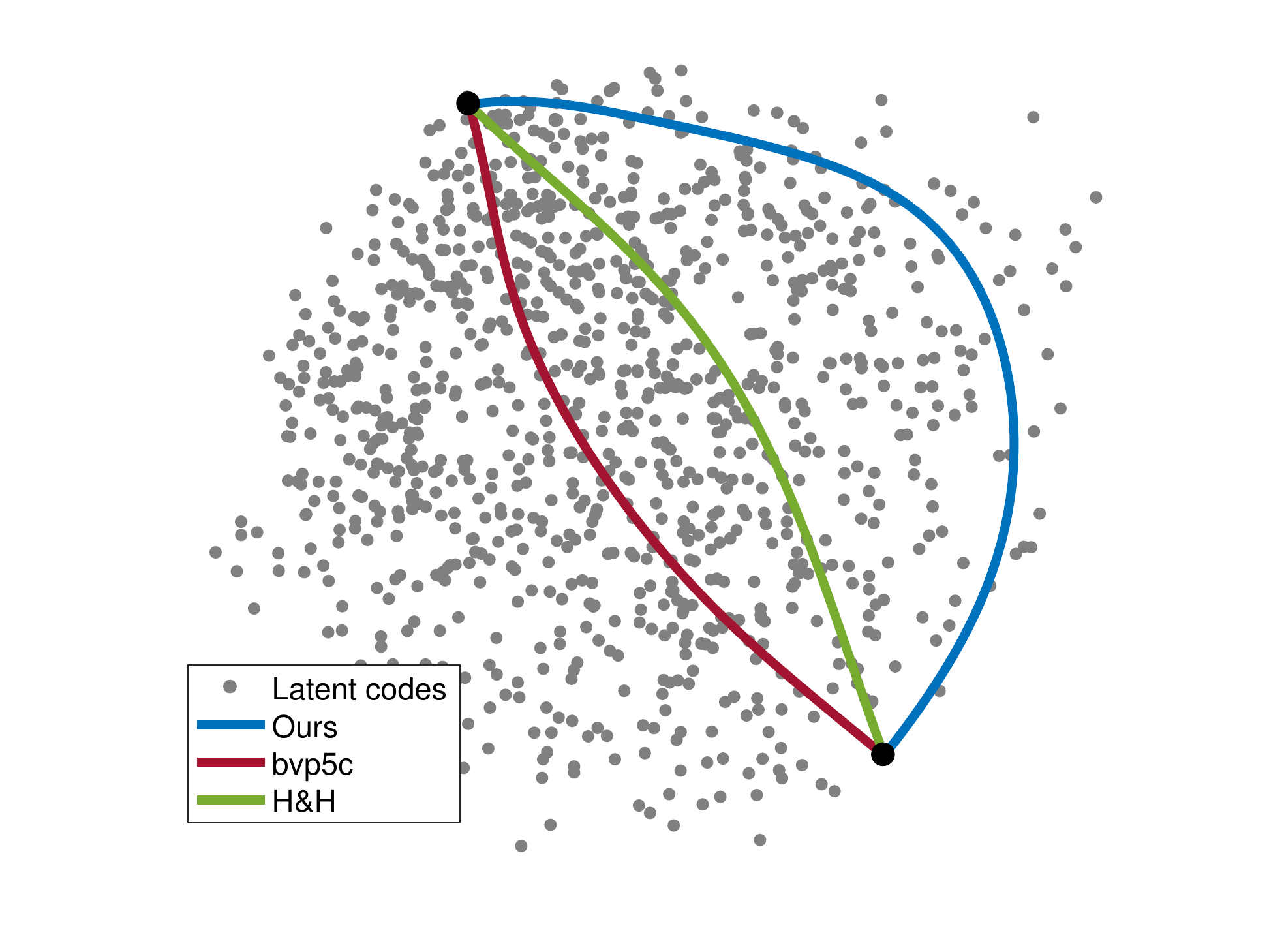}
        \caption{Latent space}
        \label{fig:ex7_fig1}
    \end{subfigure}
    ~ 
    \begin{subfigure}[b]{0.31\textwidth}
        \includegraphics[width=\textwidth]{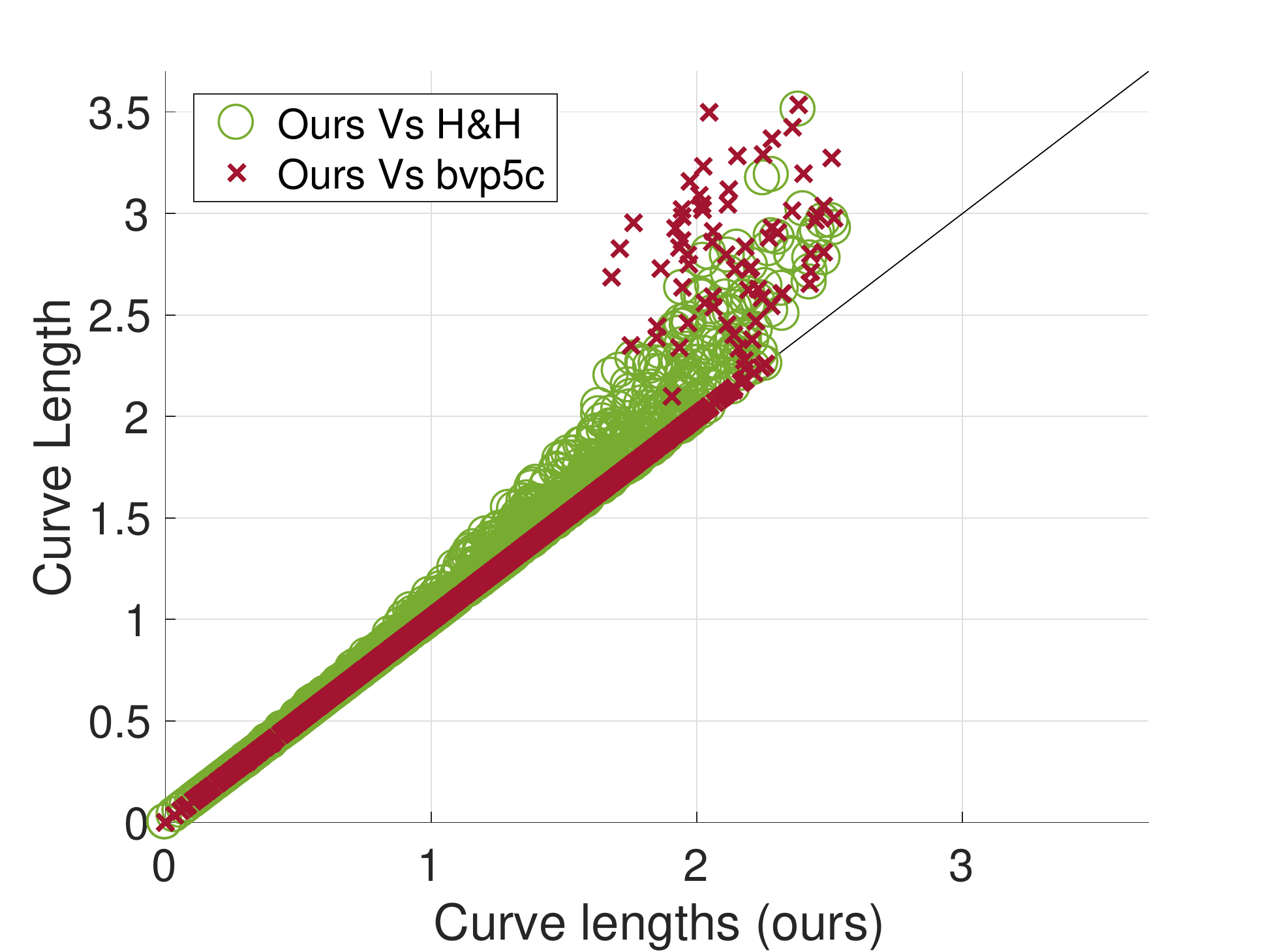}
        \caption{Curve Lengths}
        \label{fig:ex7_fig2}
    \end{subfigure}
    ~ 
    \begin{subfigure}[b]{0.31\textwidth}
        \includegraphics[width=\textwidth]{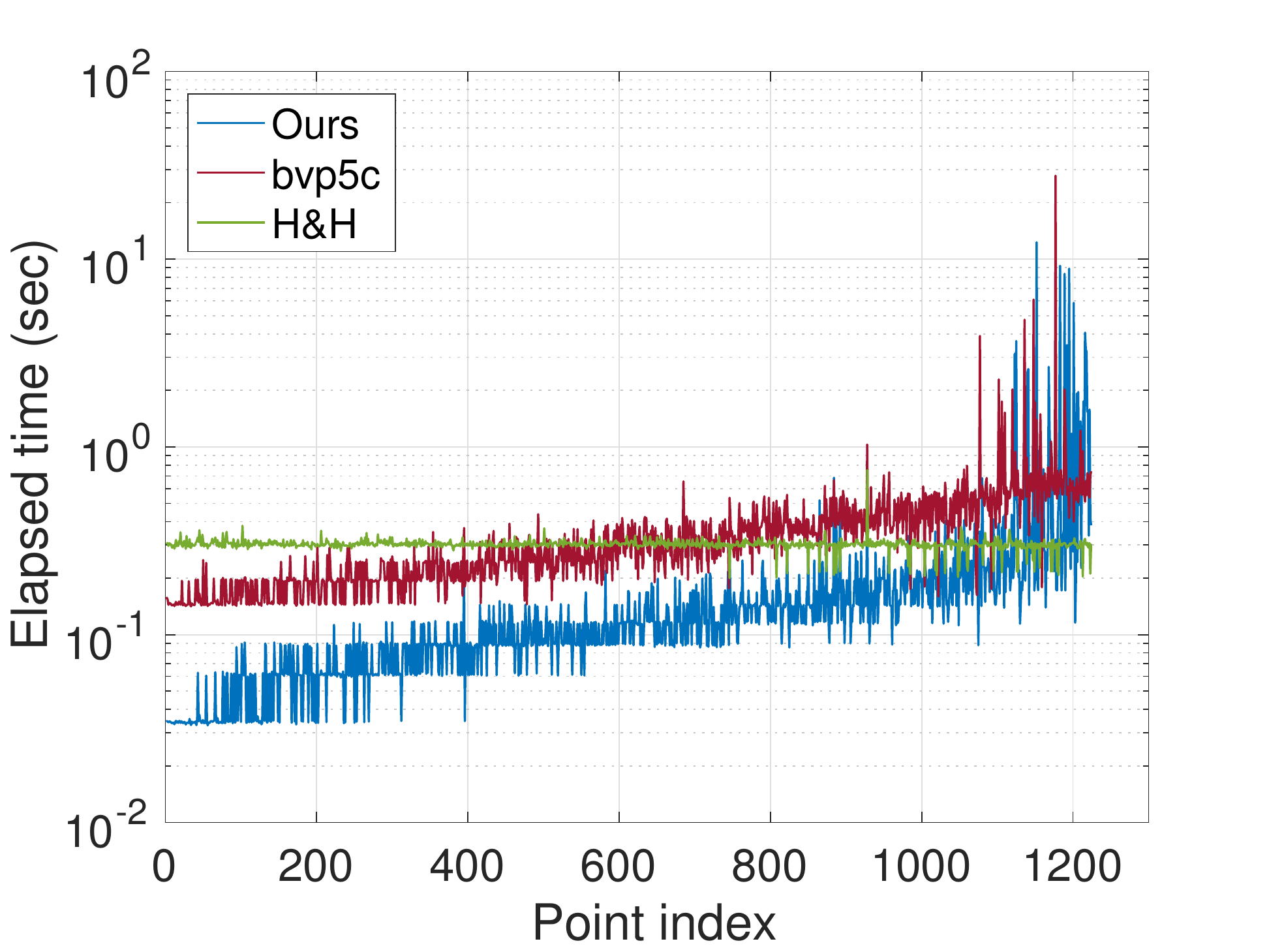}
        \caption{Runtimes}
        \label{fig:ex7_fig3}
    \end{subfigure}
    \caption{\emph{Left}: The latent space together with the computed geodesics between two points. \emph{Middle}: The curve lengths, on the $x$ axis we show the curve length of the proposed model. The results are sorted with respect to our model. \emph{Right}: The corresponding runtimes.}\label{fig:ex7}
\end{figure*}

The Variational Auto-Encoder (VAE) \citep{kingma:iclr:2014, rezende:icml:2014}, provides a systematic way to learn a low dimensional latent representation of the data, together with a generator that learns to interpolate the data manifold in the input space. Usually, deep neural networks are used to model the generator. These flexible models are able to compensate for any reparametrization of the latent space, which renders the latent space unidendifiable. Recently, \cite{arvanitidis:iclr:2018} defined a Riemannian metric in the latent space, which is induced by the generator and is invariant to the parametrization of the latent space. This resolves the identifiability issue and makes computations in the latent space parametrization invariant. More specifically, the VAEs utilizes a stochastic generator that maps 
a point $\b{x}$ from the latent space $\mathcal{X}$ to a point $\b{y}$ in the input space $\mathcal{Y}$, and it consists of two parts: the mean and the variance function as $\b{y}(\b{x}) = \bs{\mu}(\b{x}) + \bs{\sigma}(\b{x}) \odot \epsilon$, where $\epsilon\sim\mathcal{N}(0,\mathbb{I}_{\text{dim}(\mathcal{Y})})$ and $\odot$ is the pointwise multiplication. This stochastic mapping introduces a random Riemannian metric in the latent space. However, as it is shown \citep{Tosi:UAI:2014} we are able to use the expectation of the metric which has the appealing form
\begin{align}
\b{M}(\b{x}) = \b{J}_{\bs{\mu}}(\b{x})^{\Trans} \b{J}_{\bs{\mu}}(\b{x}) +  \b{J}_{\bs{\sigma}}(\b{x})^{\Trans} \b{J}_{\bs{\sigma}}(\b{x}),
\end{align}
where $\b{J}$ stands for the Jacobian of the corresponding functions. The interpretation of the metric is relatively simple. It represents the distortions due to the mean function and the uncertainty of the generator.

In this context, for the data in the input space we generated the upper half of a 2-dim sphere in $\mathbb{R}^3$, added Gaussian noise $\mathcal{N}(0,0.01)$ and scaled the data in the interval $[-1,1]^3$. Then, we trained a VAE, using for the generator a simple deep network consisted of two hidden layers with 16 units per layer, the $\texttt{softplus}$ as activation functions, and $\texttt{tanh}$ for the output layer. We used a 2-dimensional latent space, and the encoded data can be seen in Fig. \ref{fig:ex7_fig1}. There, we show the computed geodesic for the 3 methods. Interestingly, we see that the 3 resulting curves differ, and the estimated curve lengths are: 
proposed (2.52), \bvpfc{} (3.65) and H{\&}H (3.30). Our model, manages to find the shortest path, which is a particularly curved path but the second derivative remains relatively smooth, while \bvpfc{} finds a simpler curve with larger length. This is not surprising since \bvpfc{} prefers solutions where the curve is smoother, while our method prefers curves where the second derivative is smoother. In order to further analyse this behavior, we randomly pick 50 points and compute all pairwise distances. The results in Fig. \ref{fig:ex7_fig2} shows that the proposed method manages to find always the shortest path, while the other methods when the distances increase, provide a suboptimal solution. Comparing the runtimes (Fig. \ref{fig:ex7_fig3}) we see that our method is faster in the simple problems, and has only a small overhead in the difficult problems, however, it manages always to find the shortest path.

\begin{figure}[ht]
  \begin{center}
    \includegraphics[width=0.5\textwidth]{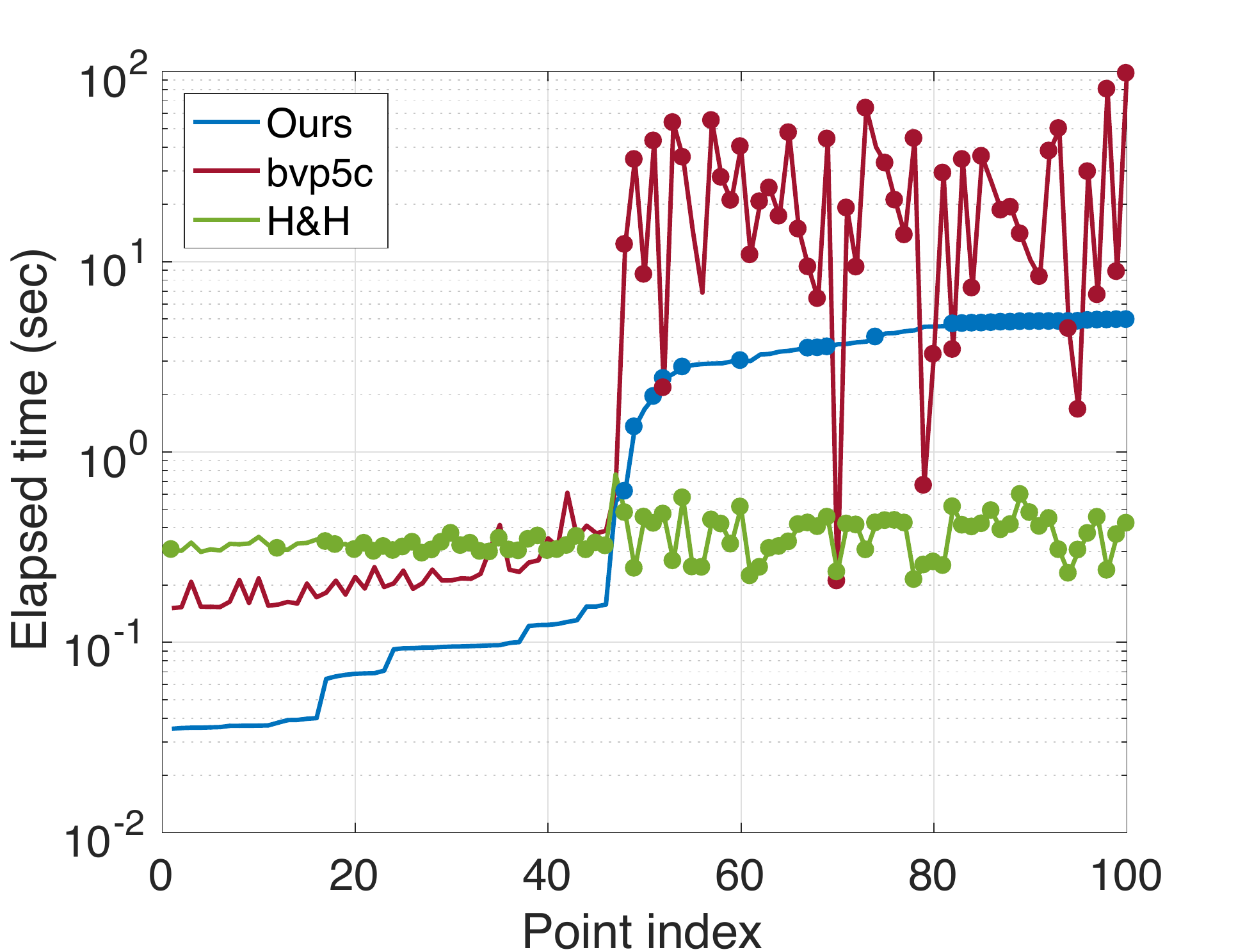}
  \end{center}
  \vspace{-10pt}
  \caption{Runtime comparison}
  \label{fig:ex6}
\end{figure}
As a last experiment, we generated a 2-dimensional sphere in $\mathbb{R}^3$ and moved the upper half by 1, and again added Gaussian noise and scaled the data to $[-1,1]^3$. Instead of $\texttt{softplus}$ in the hidden layers, we used the $\texttt{tanh}$ activation functions which increases the curvature. Here the latent space is 2-dimensional. We fix randomly a point in the latent space and compute the geodesic to 100 randomly chosen points. As we observe from the results in Fig. \ref{fig:ex6}, as long as we compute the distance between points of the same semi-sphere the runtimes of our method and \bvpfc{} are comparable. However, when the points belong in different semi-spheres the runtimes increase significantly. The reason is that curvature increases dramatically when we cross parts of the latent space where the generator is uncertain. That is also the reason why many problems are not solved (dots in the figure), but even in this challenging setting our model is more robust.


\section{Conclusions}

We proposed a simple, fast and robust algorithm to compute 
shortest paths on Riemannian manifolds learned from data. Here, standard solvers often fail due to ill-conditioned Jacobians
of the associated BVP.
Instead, our method applies a Jacobian-free fixed-point iterative scheme. The assumption is that the true path can be approximated smoothly by the predictive GP posterior. This solver makes the Riemannian methods more feasible since robustly in reasonable time, complex statistical models \citep{arvanitidis:nips:2016} can be fitted,
as well as distances can be computed in challenging deep metric scenarios \citep{arvanitidis:iclr:2018}.
 

This has been achieved by analyzing and extending
an existing probabilistic numerical solver \citep{HennigAISTATS2014},
turning it from a proof-of-concept into a
principled algorithm.
The presented method thus contributes both to
Riemannian methods and to probabilistic numerics.
Further improvements might be achieved with
more complex fixed-point iterations \citep{ishikawa1974fixed},
advanced line searches \citep{nips2015_5753},
adaptive mesh selection \citep{mazzia2004hybrid},
and improved model selection \citep{vaart2011information}.

%
%



\clearpage

\subsubsection*{Acknowledgments}

GA is supported by the Danish Center for Big Data Analytics Driven Innovation. SH was supported by a research grant (15334) from VILLUM FONDEN. This project has received funding from the European Research Council (ERC) under the European Union’s Horizon 2020 research and innovation programme (grant agreement n\textsuperscript{o} 757360). We gratefully acknowledge the support of the NVIDIA Corporation with the donation of the used Titan Xp GPU. PH gratefully acknowledges financial support by the ERC StC Action 757275 PANAMA and grant BMBF 01 IS 18 052-B of the German Federal Ministry for Education and Research.

\small

\bibliographystyle{abbrvnat}
\bibliography{bib-paper.bib}

\begin{thebibliography}{41}
\providecommand{\natexlab}[1]{#1}
\providecommand{\url}[1]{\texttt{#1}}
\expandafter\ifx\csname urlstyle\endcsname\relax
  \providecommand{\doi}[1]{doi: #1}\else
  \providecommand{\doi}{doi: \begingroup \urlstyle{rm}\Url}\fi

\bibitem[Arvanitidis et~al.(2016)Arvanitidis, Hansen, and
  Hauberg]{arvanitidis:nips:2016}
G.~Arvanitidis, L.~K. Hansen, and S.~Hauberg.
\newblock A locally adaptive normal distribution.
\newblock In \emph{Advances in Neural Information Processing Systems (NIPS)},
  2016.

\bibitem[Arvanitidis et~al.(2018)Arvanitidis, Hansen, and
  Hauberg]{arvanitidis:iclr:2018}
G.~Arvanitidis, L.~K. Hansen, and S.~Hauberg.
\newblock Latent space oddity: on the curvature of deep generative models.
\newblock In \emph{International Conference on Learning Representations
  (ICLR)}, 2018.

\bibitem[Ascher et~al.(1994)Ascher, Mattheij, and Russell]{ascher1994numerical}
U.~M. Ascher, R.~M. Mattheij, and R.~D. Russell.
\newblock \emph{Numerical solution of boundary value problems for ordinary
  differential equations}, volume~13.
\newblock Siam, 1994.

\bibitem[Bello et~al.(2017)Bello, Alkali, and Roko]{BELLO2017}
N.~Bello, A.~J. Alkali, and A.~Roko.
\newblock A fixed point iterative method for the solution of two-point boundary
  value problems for a second order differential equations.
\newblock \emph{Alexandria Engineering Journal}, 2017.
\newblock \doi{10.1016/j.aej.2017.09.010}.

\bibitem[Briol et~al.(2015)Briol, Oates, Girolami, and Osborne]{briol2015frank}
F.-X. Briol, C.~Oates, M.~Girolami, and M.~A. Osborne.
\newblock Frank-wolfe bayesian quadrature: Probabilistic integration with
  theoretical guarantees.
\newblock In \emph{Advances in Neural Information Processing Systems}, pages
  1162--1170, 2015.

\bibitem[Chkrebtii et~al.(2016)Chkrebtii, Campbell, Calderhead, and
  Girolami]{chkrebtii2016bayesian}
O.~A. Chkrebtii, D.~A. Campbell, B.~Calderhead, and M.~A. Girolami.
\newblock Bayesian solution uncertainty quantification for differential
  equations.
\newblock \emph{Bayesian Anal.}, 11\penalty0 (4):\penalty0 1239--1267, 12 2016.
\newblock \doi{10.1214/16-BA1017}.

\bibitem[Cockayne et~al.(2016)Cockayne, Oates, Sullivan, and
  Girolami]{cockayne2016probabilistic}
J.~Cockayne, C.~Oates, T.~Sullivan, and M.~Girolami.
\newblock Probabilistic meshless methods for partial differential equations and
  bayesian inverse problems.
\newblock \emph{arXiv preprint arXiv:1605.07811}, 2016.

\bibitem[Deuflhard(2011)]{deuflhard2011newton}
P.~Deuflhard.
\newblock \emph{Newton methods for nonlinear problems: affine invariance and
  adaptive algorithms}.
\newblock Springer Science \& Business Media, 2011.

\bibitem[do~Carmo(1992)]{docarmo:1992}
M.~do~Carmo.
\newblock \emph{Riemannian Geometry}.
\newblock Mathematics (Boston, Mass.). Birkh{\"a}user, 1992.

\bibitem[Feragen et~al.(2015)Feragen, Lauze, and Hauberg]{feragen2015geodesic}
A.~Feragen, F.~Lauze, and S.~Hauberg.
\newblock Geodesic exponential kernels: When curvature and linearity conflict.
\newblock In \emph{Proceedings of the IEEE Conference on Computer Vision and
  Pattern Recognition}, pages 3032--3042, 2015.

\bibitem[Hauberg et~al.(2012)Hauberg, Freifeld, and Black]{hauberg:nips:2012}
S.~Hauberg, O.~Freifeld, and M.~Black.
\newblock {A Geometric Take on Metric Learning}.
\newblock In P.~Bartlett, F.~Pereira, C.~Burges, L.~Bottou, and K.~Weinberger,
  editors, \emph{{Advances in Neural Information Processing Systems (NIPS)}},
  pages 2033--2041. MIT Press, 2012.

\bibitem[Hauberg et~al.(2015)Hauberg, Schober, Liptrot, Hennig, and
  Feragen]{hauberg2015random}
S.~Hauberg, M.~Schober, M.~Liptrot, P.~Hennig, and A.~Feragen.
\newblock A random riemannian metric for probabilistic shortest-path
  tractography.
\newblock In \emph{Medical Image Computing and Computer-Assisted
  Intervention--MICCAI 2015}, volume~18, Munich, Germany, Sept. 2015. Springer.

\bibitem[He(2000)]{HE2000115}
J.-H. He.
\newblock Variational iteration method for autonomous ordinary differential
  systems.
\newblock \emph{Applied Mathematics and Computation}, 114\penalty0
  (2):\penalty0 115 -- 123, 2000.
\newblock ISSN 0096-3003.
\newblock \doi{10.1016/S0096-3003(99)00104-6}.

\bibitem[Hennig and Hauberg(2014)]{HennigAISTATS2014}
P.~Hennig and S.~Hauberg.
\newblock {Probabilistic Solutions to Differential Equations and their
  Application to Riemannian Statistics}.
\newblock In \emph{{Proc. of the 17th int. Conf. on Artificial Intelligence and
  Statistics ({AISTATS})}}, volume~33. JMLR, W\&CP, 2014.

\bibitem[Hennig et~al.(2015)Hennig, Osborne, and
  Girolami]{hennig15probabilistic}
P.~Hennig, M.~A. Osborne, and M.~Girolami.
\newblock Probabilistic numerics and uncertainty in computations.
\newblock \emph{Proceedings of the Royal Society of London A: Mathematical,
  Physical and Engineering Sciences}, 471\penalty0 (2179), 2015.

\bibitem[Ishikawa(1974)]{ishikawa1974fixed}
S.~Ishikawa.
\newblock Fixed points by a new iteration method.
\newblock \emph{Proceedings of the American Mathematical Society}, 44\penalty0
  (1):\penalty0 147--150, 1974.

\bibitem[Jafari(2014)]{JAFARI20141}
H.~Jafari.
\newblock A comparison between the variational iteration method and the
  successive approximations method.
\newblock \emph{Applied Mathematics Letters}, 32:\penalty0 1 -- 5, 2014.
\newblock ISSN 0893-9659.
\newblock \doi{10.1016/j.aml.2014.02.004}.

\bibitem[Johnson(1972)]{johnson1972fixed}
G.~G. Johnson.
\newblock Fixed points by mean value iterations.
\newblock \emph{Proceedings of the American Mathematical Society}, 34\penalty0
  (1):\penalty0 193--194, 1972.

\bibitem[{Kanagawa} et~al.(2018){Kanagawa}, {Hennig}, {Sejdinovic}, and
  {Sriperumbudur}]{kanagawa2018gps}
M.~{Kanagawa}, P.~{Hennig}, D.~{Sejdinovic}, and B.~K. {Sriperumbudur}.
\newblock {Gaussian Processes and Kernel Methods: A Review on Connections and
  Equivalences}.
\newblock \emph{ArXiv e-prints}, July 2018.

\bibitem[{Kersting} et~al.(2018){Kersting}, {Sullivan}, and
  {Hennig}]{kersting2018arXiv}
H.~{Kersting}, T.~J. {Sullivan}, and P.~{Hennig}.
\newblock {Convergence Rates of Gaussian ODE Filters}.
\newblock \emph{ArXiv e-prints}, July 2018.

\bibitem[Kersting and Hennig(2016)]{KerstingHennigUAI2016}
H.~P. Kersting and P.~Hennig.
\newblock Active uncertainty calibration in {B}ayesian {ODE} solvers.
\newblock In Janzing and Ihlers, editors, \emph{Uncertainty in Artificial
  Intelligence (UAI)}, volume~32, 2016.

\bibitem[Khuri and Sayfy(2014)]{KHURI201428}
S.~Khuri and A.~Sayfy.
\newblock Variational iteration method: Green's functions and fixed point
  iterations perspective.
\newblock \emph{Applied Mathematics Letters}, 32:\penalty0 28 -- 34, 2014.
\newblock ISSN 0893-9659.
\newblock \doi{10.1016/j.aml.2014.01.006}.

\bibitem[Kimeldorf and Wahba(1970)]{kimeldorf1970correspondence}
G.~S. Kimeldorf and G.~Wahba.
\newblock A correspondence between bayesian estimation on stochastic processes
  and smoothing by splines.
\newblock \emph{The Annals of Mathematical Statistics}, 41\penalty0
  (2):\penalty0 495--502, 1970.

\bibitem[Kingma and Welling(2014)]{kingma:iclr:2014}
D.~P. Kingma and M.~Welling.
\newblock Auto-{E}ncoding {V}ariational {B}ayes.
\newblock In \emph{Proceedings of the 2nd International Conference on Learning
  Representations (ICLR)}, 2014.

\bibitem[Mahsereci and Hennig(2015)]{nips2015_5753}
M.~Mahsereci and P.~Hennig.
\newblock Probabilistic line searches for stochastic optimization.
\newblock In \emph{Advances in Neural Information Processing Systems}, pages
  181--189, 2015.

\bibitem[Mann(1953)]{mann1953mean}
W.~R. Mann.
\newblock Mean value methods in iteration.
\newblock \emph{Proceedings of the American Mathematical Society}, 4\penalty0
  (3):\penalty0 506--510, 1953.

\bibitem[Mazzia and Trigiante(2004)]{mazzia2004hybrid}
F.~Mazzia and D.~Trigiante.
\newblock A hybrid mesh selection strategy based on conditioning for boundary
  value ode problems.
\newblock \emph{Numerical Algorithms}, 36\penalty0 (2):\penalty0 169--187,
  2004.

\bibitem[Micchelli et~al.(2006)Micchelli, Xu, and
  Zhang]{micchelli2006universal}
C.~A. Micchelli, Y.~Xu, and H.~Zhang.
\newblock Universal kernels.
\newblock \emph{Journal of Machine Learning Research}, 7\penalty0
  (Dec):\penalty0 2651--2667, 2006.

\bibitem[Oates et~al.(2017)Oates, Cockayne, Aykroyd, et~al.]{oates2017bayesian}
C.~Oates, J.~Cockayne, R.~G. Aykroyd, et~al.
\newblock Bayesian probabilistic numerical methods for industrial process
  monitoring.
\newblock \emph{arXiv preprint arXiv:1707.06107}, 2017.

\bibitem[Picard(1890)]{picard1890}
E.~Picard.
\newblock M\'emoire sur la th\'eorie des \'equations aux d\'eriv\'ees
  partielles et la m\'ethode des approximations successives.
\newblock \emph{Journal de Math\'ematiques Pures et Appliqu\'es}, 6:\penalty0
  145--210, 1890.

\bibitem[Rasmussen and Williams(2006)]{rasmussenwilliams}
C.~Rasmussen and C.~Williams.
\newblock \emph{{Gaussian Processes for Machine Learning}}.
\newblock MIT, 2006.

\bibitem[Rezende et~al.(2014)Rezende, Mohamed, and Wierstra]{rezende:icml:2014}
D.~J. Rezende, S.~Mohamed, and D.~Wierstra.
\newblock Stochastic backpropagation and approximate inference in deep
  generative models.
\newblock In \emph{Proceedings of the 31st International Conference on Machine
  Learning}, Bejing, China, 2014.

\bibitem[{Schober} et~al.(2014){Schober}, {Duvenaud}, and
  {Hennig}]{schober2014nips}
M.~{Schober}, D.~{Duvenaud}, and P.~{Hennig}.
\newblock {Probabilistic ODE Solvers with Runge-Kutta Means}.
\newblock \emph{{Advances in Neural Information Processing Systems (NIPS)}},
  2014.

\bibitem[Schober et~al.(2018)Schober, S{\"a}rkk{\"a}, and
  Hennig]{schober2017probabilistic}
M.~Schober, S.~S{\"a}rkk{\"a}, and P.~Hennig.
\newblock A probabilistic model for the numerical solution of initial value
  problems.
\newblock \emph{Statistics and Computing}, Jan 2018.
\newblock \doi{10.1007/s11222-017-9798-7}.

\bibitem[Skilling(1991)]{skilling1991bayesian}
J.~Skilling.
\newblock {Bayesian solution of ordinary differential equations}.
\newblock \emph{Maximum Entropy and Bayesian Methods, Seattle}, 1991.

\bibitem[Teymur et~al.(2016)Teymur, Zygalakis, and
  Calderhead]{teymur2016probabilistic}
O.~Teymur, K.~Zygalakis, and B.~Calderhead.
\newblock Probabilistic linear multistep methods.
\newblock In D.~D. Lee, M.~Sugiyama, U.~V. Luxburg, I.~Guyon, and R.~Garnett,
  editors, \emph{Advances in Neural Information Processing Systems 29}, pages
  4314--4321. Curran Associates, Inc., 2016.

\bibitem[Tosi et~al.(2014)Tosi, Hauberg, Vellido, and Lawrence]{Tosi:UAI:2014}
A.~Tosi, S.~Hauberg, A.~Vellido, and N.~D. Lawrence.
\newblock {Metrics for Probabilistic Geometries}.
\newblock In \emph{The Conference on Uncertainty in Artificial Intelligence
  (UAI)}, July 2014.

\bibitem[Vaart and Zanten(2011)]{vaart2011information}
A.~v.~d. Vaart and H.~v. Zanten.
\newblock Information rates of nonparametric gaussian process methods.
\newblock \emph{Journal of Machine Learning Research}, 12\penalty0
  (Jun):\penalty0 2095--2119, 2011.

\bibitem[Wahba(1990)]{wahba1990spline}
G.~Wahba.
\newblock \emph{{Spline models for observational data}}.
\newblock Number~59 in {CBMS-NSF Regional Conferences series in applied
  mathematics}. SIAM, 1990.

\bibitem[Wendland(2004)]{wendland2004scattered}
H.~Wendland.
\newblock \emph{Scattered data approximation}.
\newblock Cambridge University Press, 2004.

\bibitem[Xi et~al.(2018)Xi, Briol, and Girolami]{briol2018bqmultiple}
X.~Xi, F.-X. Briol, and M.~Girolami.
\newblock {Bayesian quadrature for multiple related integrals}.
\newblock In \emph{International Conference on Machine Learning}, 2018.

\end{thebibliography}

\clearpage
\appendix

%
%
%
%
%

\section{Approximate Shortest Paths}
\label{sec:app:approximate_sp}

The proposed approximation to the shortest path is the posterior mean of a Gaussian process, and is parametrized by a set of second derivatives $\ddot{\b{z}}_n$ on a discrete mesh $\Delta = \{t_0=0, t_1, \dots, t_{N-1}=1\} \subset [0,1]$ of evaluation knots $t_n$. Therefore, the shortest path is
\begin{align}
&\bs{\mu}(t) = \b{m}(t) 
+ \bs{\omega}\Trans \text{vec}\left(
\begin{bmatrix}\b{x} - \b{m}(0)\\ \b{y} - \b{m}(1)\\\ddot{\b{z}}_\mesh - \b{\ddot{m}}(\mesh)\end{bmatrix}\Trans \right)\nonumber\\
&\b{G} = \b{V} \otimes \bigg(\begin{bmatrix}k(\bmesh,\bmesh) & \frac{\partial^2}{\partial s^2}k(\bmesh, \mesh)\\ \frac{\partial^2}{\partial t^2}k(\mesh,\bmesh) &  \frac{\partial^4}{\partial t^2\partial s^2}k(\mesh,\mesh)\end{bmatrix}\\
&\phantom{\b{G} = \b{V} \otimes \bigg(}
+ \operatorname{diag}(0,0,\b{\Sigma},\dotsc,\b{\Sigma})\bigg)\nonumber\\
&\bs{\omega}\Trans = \left(\b{V} \otimes
\begin{bmatrix}k(t,\bmesh) & \frac{\partial^2}{\partial s^2}k(t,\mesh)\end{bmatrix}
\right) \b{G}^{-1}\nonumber.
\end{align}
A fixed-point scheme to learn the parameters $\ddot{\b{z}}_\mesh$ that satisfy the ODE of the geodesic curve is presented in Sec.~\ref{sec:method_description}. Next we show how the components of the GP can be chosen.

\textbf{Mean function}

The most natural choice regarding the mean function of the prior is the straight line that connects the two boundary points $\b{m}(t) = \b{c}(0)\cdot t + (\b{c}(1) - \b{c}(0)) \cdot (1-t)$. This is the shortest path when the Riemannian manifold is flat. Also, when the curvature of the manifold is low, then the shortest path will be relatively close to the straight line. Likewise, when two points are very close on the manifold. Note that the mean function of the prior is the initial guess of the BVP solution.

For instance, if for the kernel we chose the SE, then implicitly the prior assumption is that the shortest paths are
smooth curves varying on a length scale of $\lambda$ along $t$. Also, the amplitude $\b{V} = [(\b{a} - \b{b})\Trans \b{S}_\b{x} (\b{a} - \b{b})]\cdot \b{S}_\b{x} \in \mathbb{R}^{D\times D}$, where $\b{S}_\b{x}$ is the
sample covariance of the dataset $\b{x}_{1:N}$ as in \cite{HennigAISTATS2014}.

\textbf{Kernel}

The kernel type implies the smoothness of the approximated
curve. Since shortest paths are expected to be relatively smooth as two times
differentiable parametric functions, a reasonable choice for the kernel is to be
smooth, e.g. squared exponential (SE), Matern, etc.

Moreover, it is important
to use \emph{stationary} kernels, since they treat the two boundary points equally.
For example, the non-stationary Wienner kernel is a common choice for IVPs.
However, in a BVP such a kernel is a poor fit, because if the time
interval is inverted, then the resulting curve will be different.

\textbf{Mesh}

The \emph{Reproducing Kernel Hilbert Space (RKHS)} \citep{rasmussenwilliams} of the Gaussian process is spanned by the basis functions $\{k(t_n, t)\}_{n=0}^{N-1}$. The predictive posterior $\bs{\mu}(t)$ lies in the RKHS as a linear combination of the basis functions $k(t_n, t)$. Therefore, for our approximation to work, we need the true shortest path to be approximated sufficiently well by the RKHS. This, means that the $\bs{\mu}(t)$ has to be a smooth approximation to the true path.

In our case, the mesh $\Delta$ specifies the locations, as well as the number of the basis functions. Consequently, by increasing the size of mesh, we essentially increase the RKHS such that to be able to approximate more complicated true shortest paths. However, knowing in prior the correct number and the placements of the knots is unrealistic. For that reason a reasonable solution is to use a uniform grid for the interval $[0,1]$. Moreover, $\Delta$ can be seen as a common hyper-parameter for every choice of kernel.

\textbf{Hyper-parameters}

The hyper-parameters of each kernel are kept fixed, because learning the hyper-parameters in parallel with the artificial dataset $\ddot{\b{z}}_\mesh$ may lead to degenerate solutions.

\begin{table*}[!h]
	\begin{adjustbox}{max width=\textwidth}
	\centering
	\begin{tabular}{c | c c c c c c}
		$N$ & 5 & 10 & 15  & 25 & 50 & 100\\\hline
		\#1 & 2.52($\pm$ 0.4693
		) & 2.51($\pm$ 0.3296) & 2.51($\pm$ 0.1562) & 2.49($\pm$ 0.1476) & 2.47($\pm$ 0.0043) & 2.47($\pm$ 0.0004) \\
		\#2 & 2.36($\pm$ 0.4541
		) & 2.33($\pm$ 0.1800) & 2.34($\pm$ 0.2426) & 2.32($\pm$ 0.1162) & 2.32($\pm$ 0.0011) & 2.32($\pm$ 0.0004) \\
		\#3 & 2.20($\pm$ 0.5315
		) & 2.19($\pm$ 0.1653) & 2.18($\pm$ 0.0742) & 2.17($\pm$ 0.0559) & 2.17($\pm$ 0.0017) & 2.17($\pm$ 0.0004) \\
		\#4 & 2.17($\pm$ 0.5496
		) & 2.16($\pm$ 0.1972) & 2.15($\pm$ 0.1028) & 2.15($\pm$ 0.0417) & 2.14($\pm$ 0.0020) & 2.14($\pm$ 0.0003) \\
	\end{tabular}
	\end{adjustbox}
\caption{Experiment for constant speed curves and different mesh sizes.}\label{tab:constant_speed}
\end{table*}

\section{Scaling of the algorithm with respect to mesh and dimensions}
\begin{figure}[!h]
	\centering
	\includegraphics[width=0.45\textwidth]{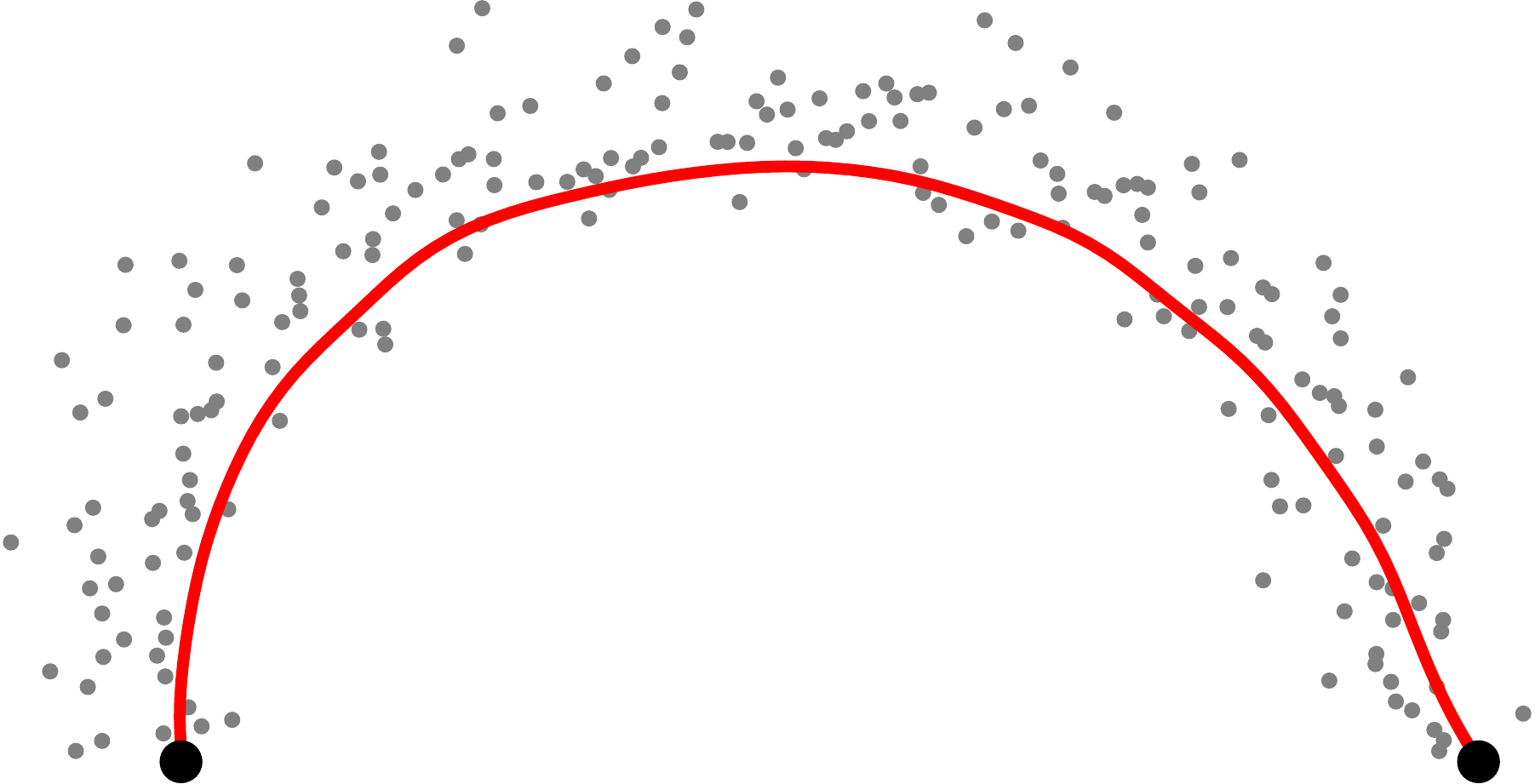}
	\caption{Example of a shortest path.}
	\label{fig:geod_example}
\end{figure}

The curvature of the Riemannian manifold $\M$ i.e., the behavior of the learned metric, implies the complexity of the shortest paths. As regards the iterations that the algorithm needs in order to find the parameters which solve the ODE, these are related to the ability of the RKHS to approximate the true shortest path. In other words, when the true shortest path can be approximated easily by the RKHS, then the only few fixed point iterations are utilized.

For instance, in Fig.~\ref{fig:geod_example} we show a challenging shortest path for a non-parametric metric with $\sigma_\M=0.1$, which means that the curvature is high. When $N=10$ the RKHS is not large enough to approximate easily the true path, so $300$ iterations are needed in order for the algorithm to converge. When we increase the $N=50$ the true path can be smoothly approximated easier by the enlarged RKHS, so that only $80$ fixed point iterations are needed. When we increase the $\sigma_\M=0.15$ the curvature of $\M$ reduces, so now $85$ and $32$ iterations are needed, respectively.

\begin{figure}[!h]
	\centering
	\includegraphics[width=0.55\textwidth]{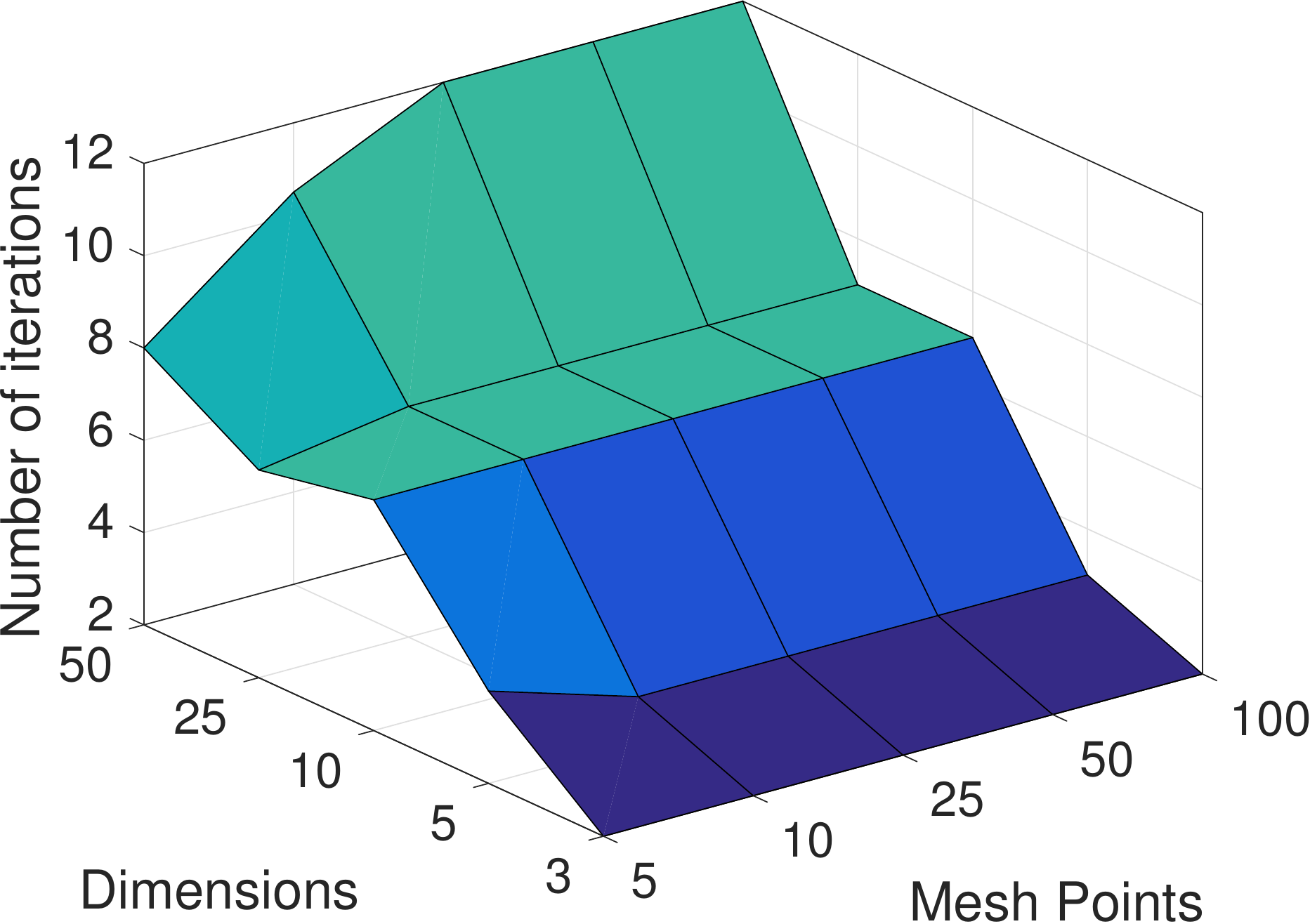}
	\caption{Scaling of the algorithm.}
	\label{fig:scaling_N_D}
\end{figure}

For completeness, we test how the method scales to higher dimensions as well. In this dataset, we fix a random point as the base point, and a subset of 100 points. We fix the $\sigma_\M=0.25$ and we chose the dimensions $[3, 5, 10, 25, 50]$ and the mesh sizes $[5, 10, 25, 50, 100]$. Then, we map the 2-dimensional dataset into each dimension using an orthogonal map, we add noise $\N(0,0.01)$ and we compute all the shortest paths between the subset and the base point for different mesh sizes. As we see in Fig.~\ref{fig:scaling_N_D}, the scaling is sublinear as regards the mesh size. Of course, as the dimension increases the problem becomes more complex, so more iterations are needed. Note that the $\M$ does not have high curvature, which means that the true shortest path can be approximated relatively easy by each RKHS.

\section{Constant Speed Curves}

The exact definition of the \emph{geodesic} is that, it is a locally minimizing curve with constant speed. This means that the geodesic might be not the global shortest path, but any segment on the geodesic curve is minimizing the length locally. However, it is important that the geodesic has constant speed. Also, by definition a curve that satisfies the ODE has constant speed.

Here we test how the mesh size $N$ affects the speed of the resulting curve. In Table~\ref{tab:constant_speed} we show the mean and the standard deviation of the speed for 4 curves in the data manifold of Fig.~\ref{fig:geod_example}. The results show that when the mesh increases, the speed becomes more constant since the standard deviation decreases. Instead, for small $N$ the curve satisfies the ODE only at the knots $t_n$, however, it does not have the exact dynamics of the true curve. In other words, with only $N$ points we are not always capable to approximate exactly the true curve. This means that our solution is a smooth approximation of the true curve, but it is not able to have constant speed. As we increase the $N$ the RKHS can approximate exactly the true curve, which satisfies the ODE for every $t$, and thus, it has constant speed.

\section{Robustness of the Solver}

\begin{figure}[!h]
\centering
\includegraphics[width=0.45\textwidth]{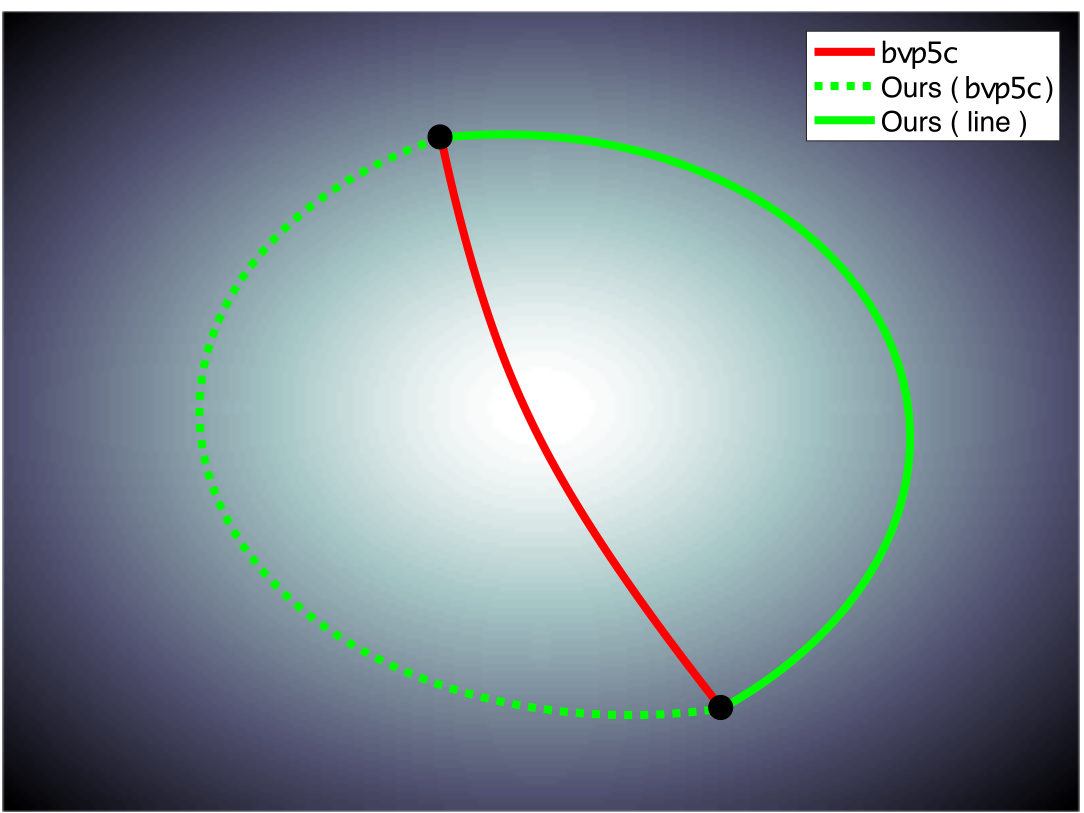}
\caption{Example of robustness.}
\label{fig:robustness}
\end{figure}

We conducted an experiment to test the robustness of our solver. In particular, we computed a challenging shortest path in the latent space of the deterministic generator $f(x,y) = [x,y,x^2 + y^2]$. In Fig.~\ref{fig:robustness} we show the paths found by \bvpfc, our method when initialized with the straight line and when it is initialized by the \bvpfc's solution. Obviously, the \bvpfc~converges to a suboptimal solution, while our method manages to find the true shortest path when initialized with the straight line. Interestingly, due to its robustness our solver manages to find a geodesic even by initializing it with the suboptimal solution of \bvpfc. Of course, this is not the shortest path but it is a geodesic, because it has constant speed as it satisfies the ODE $\forall t$, and also, it is locally length minimizing.

\section{Downstream Tasks}

We also compared the performance of our solver on downstream tasks.

From the LAND experiment (see Sec.~\ref{sec:non_parametric_experiments}) we clustered the data using the trained mixture models and a linear model, and we get the errors: 0\% (ours), 15\% (bvp5c), 21\% (linear). We also numerically measure the KL divergence between the learned distributions and the generating distribution, and observe that the proposed solver improves the fit: 0.35 (ours), 0.65 (bvp5c), 0.53 (linear).

Additionally, we performed $k$-means clustering on a 2-dimensional latent space of a VAE trained on MNIST digits 0,1,2 and the resulting errors: ~92($\pm$5)\% (ours, 1.6($\pm$0.7) hours), ~91($\pm$5)\% (bvp5c, 8($\pm$4.5) hours), ~83($\pm$4)\% (linear). The proposed model is, thus, both faster and more accurate on downstream tasks.

\end{document}